\documentclass{article}

    \PassOptionsToPackage{numbers, compress}{natbib}

\usepackage[preprint]{neurips_2024}
\usepackage{multirow}

\usepackage[utf8]{inputenc} 
\usepackage[T1]{fontenc}    
\usepackage{hyperref}       
\usepackage{url}            
\usepackage{booktabs}       
\usepackage{amsfonts}       
\usepackage{nicefrac}       
\usepackage{microtype}      
\usepackage{xcolor}         
\usepackage{graphicx}
\usepackage{svg}
\usepackage{subfigure}

\usepackage{amssymb} 
\usepackage{amsmath}

\newcommand{\ie}{\textit{i.e.}}
\newcommand{\eg}{\textit{e.g.}}
\newcommand{\etal}{\textit{et al.}}
\newcommand{\vs}{\textit{vs.}}
\newcommand{\wrt}{w.r.t.}
\newcommand{\etc}{\textit{etc.}}
\newcommand{\sota}{state-of-the-art}

\title{FaithFill: Faithful Inpainting for Object Completion Using a Single Reference Image}

\author{%
  Rupayan Mallick \\
  Department of Computer Science\\
  Georgetown University, Washington, D.C.\\
  \texttt{rupayan.mallick@georgetown.edu} \\
  \And
  Amr Abdalla \\
  Department of Computer Science\\
  Georgetown University, Washington, D.C.\\
  \texttt{aaa654@georgetown.edu} \\
  \AND
  Sarah Adel Bargal \\
  Department of Computer Science\\
  Georgetown University, Washington, D.C.\\
  \texttt{sarah.bargal@georgetown.edu} \\
}

\begin{document}

\maketitle

\begin{abstract}
  We present \textit{FaithFill}, a diffusion-based inpainting object completion approach for realistic generation of missing object parts. Typically, multiple reference images are needed to achieve such realistic generation, otherwise the generation would not faithfully preserve shape, texture, color, and background. In this work, we propose a pipeline that utilizes only a single input reference image -having varying lighting, background, object pose, and/or viewpoint. The singular reference image is used to generate multiple views of the object to be inpainted. We demonstrate that \textit{FaithFill} produces faithful generation of the object's missing parts, together with background/scene preservation, from a single reference image. This is demonstrated through standard similarity metrics, human judgement, and GPT evaluation. Our results are presented on the DreamBooth dataset, and a novel proposed dataset. 
\end{abstract}

\begin{figure*}[!h]
    \centering
    \setlength{\arrayrulewidth}{1.5pt}
    
    \rotatebox[origin=c]{0}{
    \begin{tabular}{cccccccc}
      \hspace*{2em}
      \parbox{1.5cm}{\centering \textit{Reference} \\ 
      \textit{Image}} &
      \hspace*{-0.5em}
      \parbox{1.5cm}{\centering \textit{Masked} \\ \textit{Target}} &
      \hspace*{-0.8em}
      \parbox{1.5cm}{\centering \textit{Target} \\ \textit{Image}} &
      \hspace*{0.3em}
      \parbox{1.5cm}{\centering \textit{Stable} \\ \textit{Inpainting}} &
      \hspace*{-0.2em}
      \parbox{1.5cm}{\centering \textit{Blended} \\ \textit{Latent Diffusion}} &
      \hspace*{-0.9em}
      \parbox{1.5cm}{\centering \textit{Paint By} \\ \textit{Example}} &
      \hspace*{-0.7em}
      \parbox{1.5cm}{\centering \textit{\textbf{FaithFill}} \\ \textit{\textbf{(Ours)}}}
    \end{tabular}
    }

    \rotatebox[origin=c]{90}{\parbox{1cm}{\centering\textit{Toy} \\ \textit{Monster}}\hspace*{-3.5em}} 
    \includegraphics[width=0.12\linewidth,height=0.12\linewidth,trim={0cm 0cm 0cm 0cm},clip]{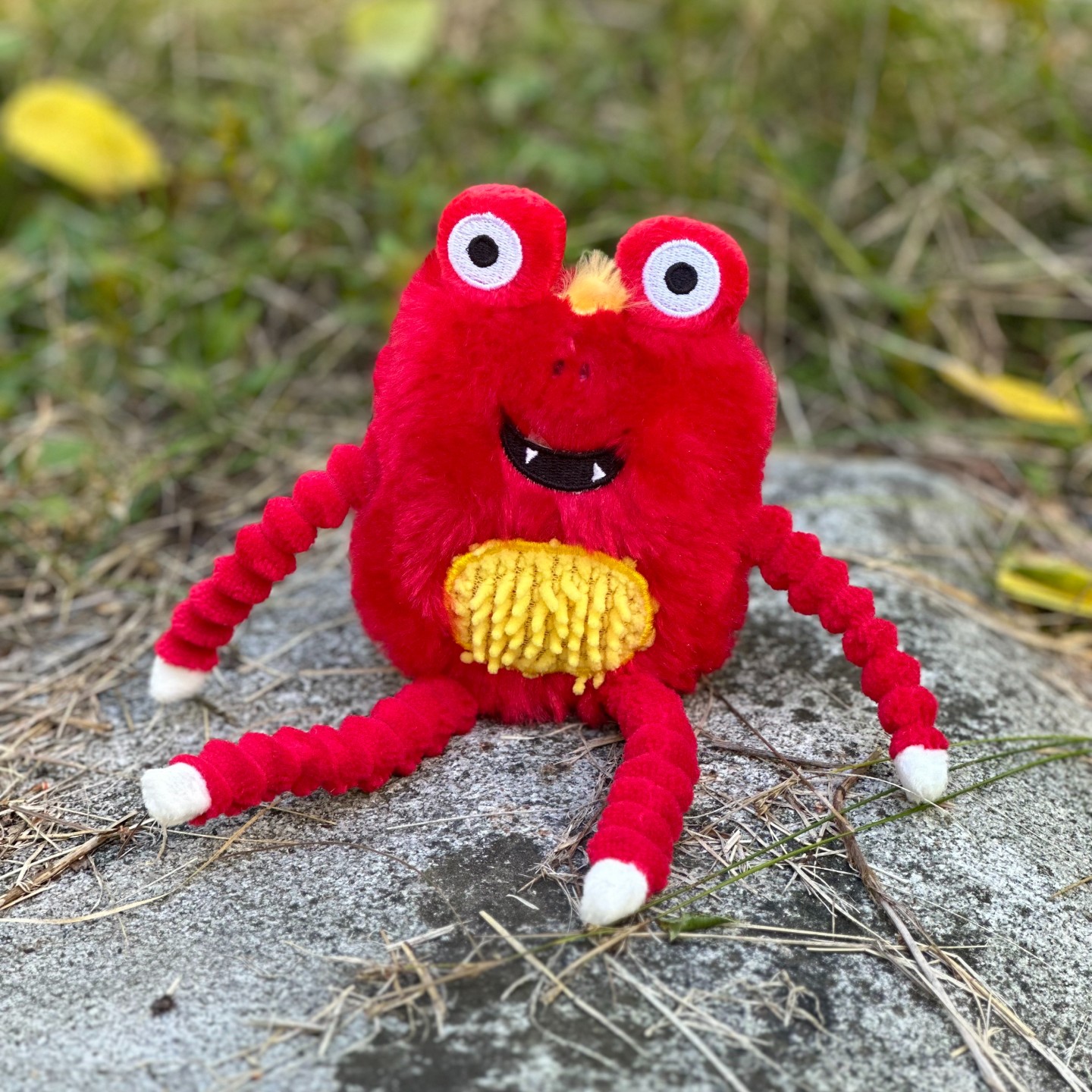} 
    \includegraphics[width=0.12\linewidth,height=0.12\linewidth,trim={0cm 0cm 0cm 0cm},clip]{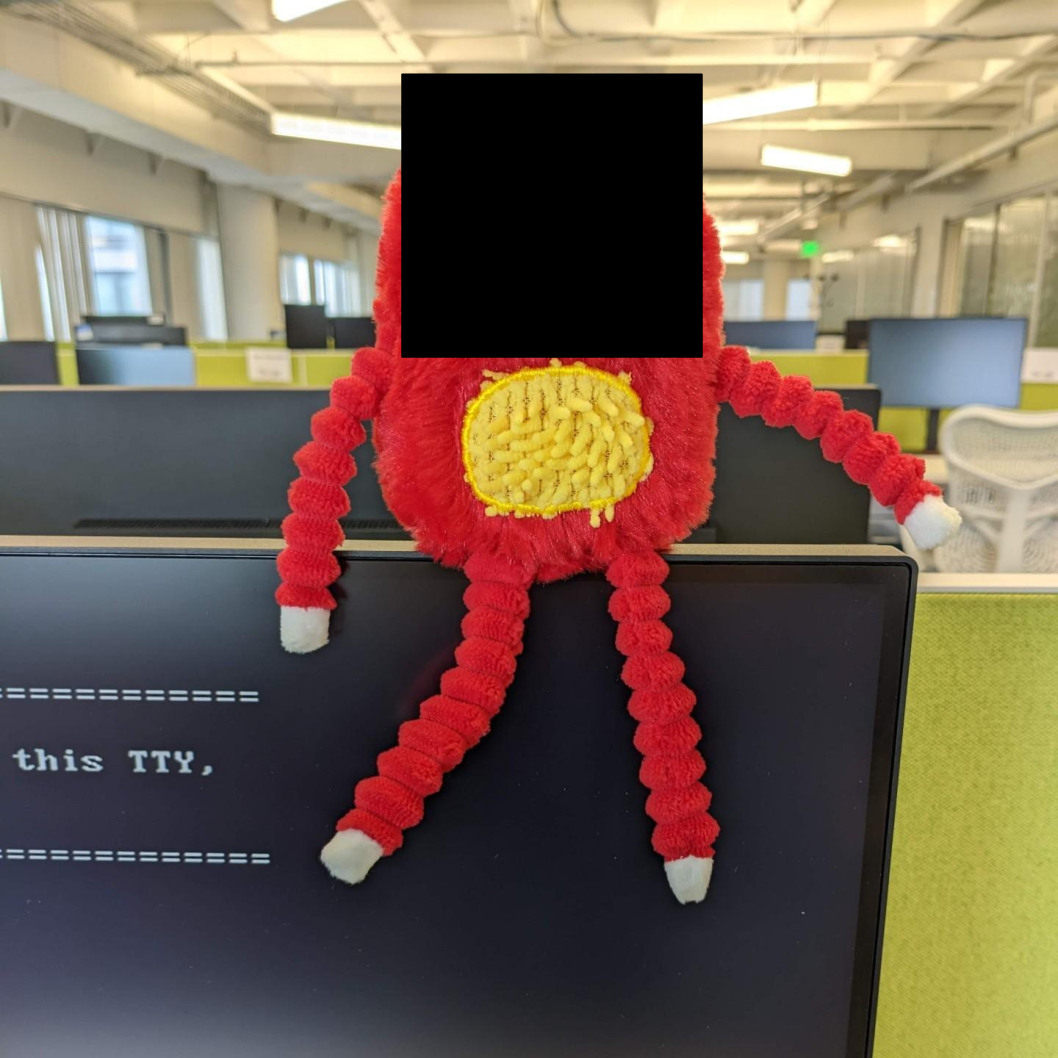}
    \includegraphics[width=0.12\linewidth,height=0.12\linewidth,trim={0cm 0cm 0cm 0cm},clip]{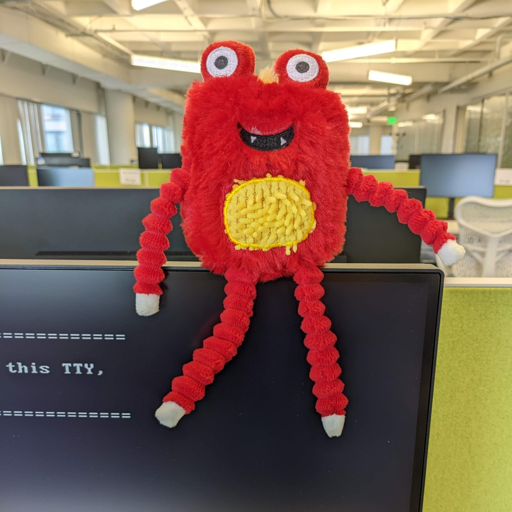} 
    \hspace*{0.1em} \vline  \hspace*{0.4em}
    \includegraphics[width=0.12\linewidth,height=0.12\linewidth,trim={0cm 0cm 0cm 0cm},clip]{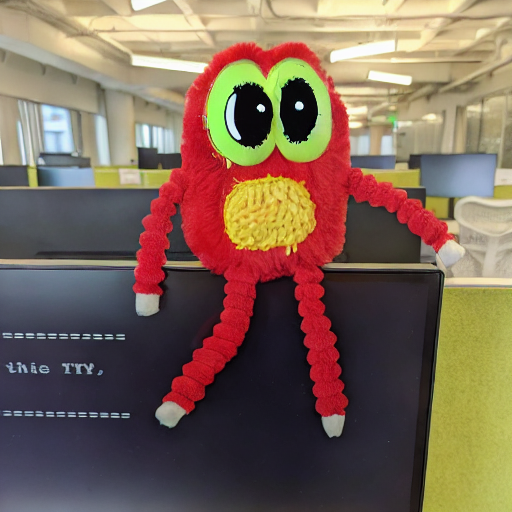}
    \includegraphics[width=0.12\linewidth,height=0.12\linewidth,trim={0cm 0cm 0cm 0cm},clip]{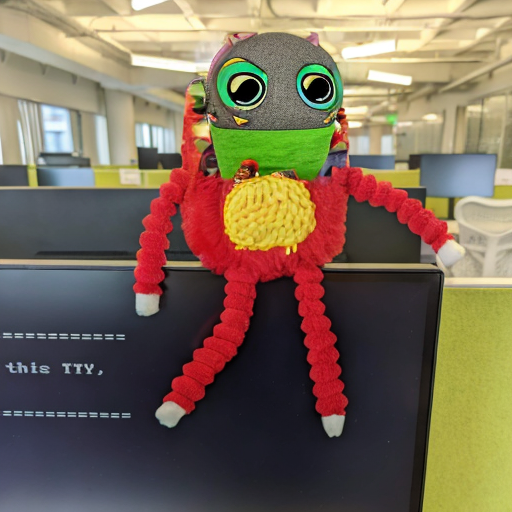}
    \includegraphics[width=0.12\linewidth,height=0.12\linewidth,trim={0cm 0cm 0cm 0cm},clip]{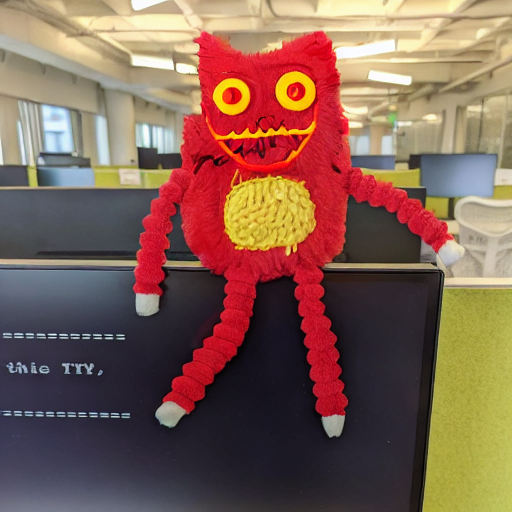}
    \includegraphics[width=0.12\linewidth,height=0.12\linewidth,trim={0cm 0cm 0cm 0cm},clip]{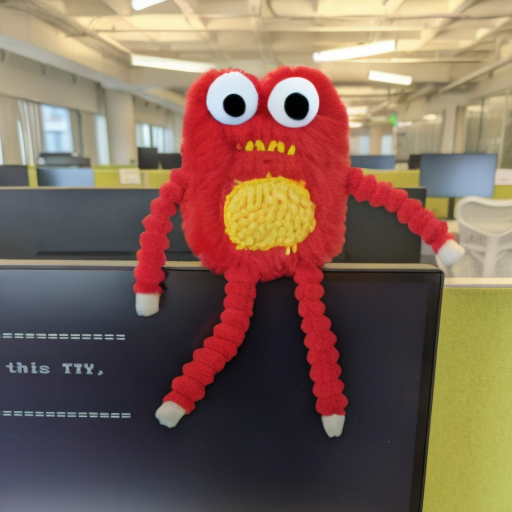} \\[0.2em]
    
    \rotatebox[origin=c]{90}{\parbox{1cm}{\centering\textit{Kitty} \\ \centering\textit{Clock}}\hspace*{-3.75em}}   
    \includegraphics[width=0.12\linewidth,height=0.12\linewidth,trim={0cm 0cm 0cm 0cm},clip]{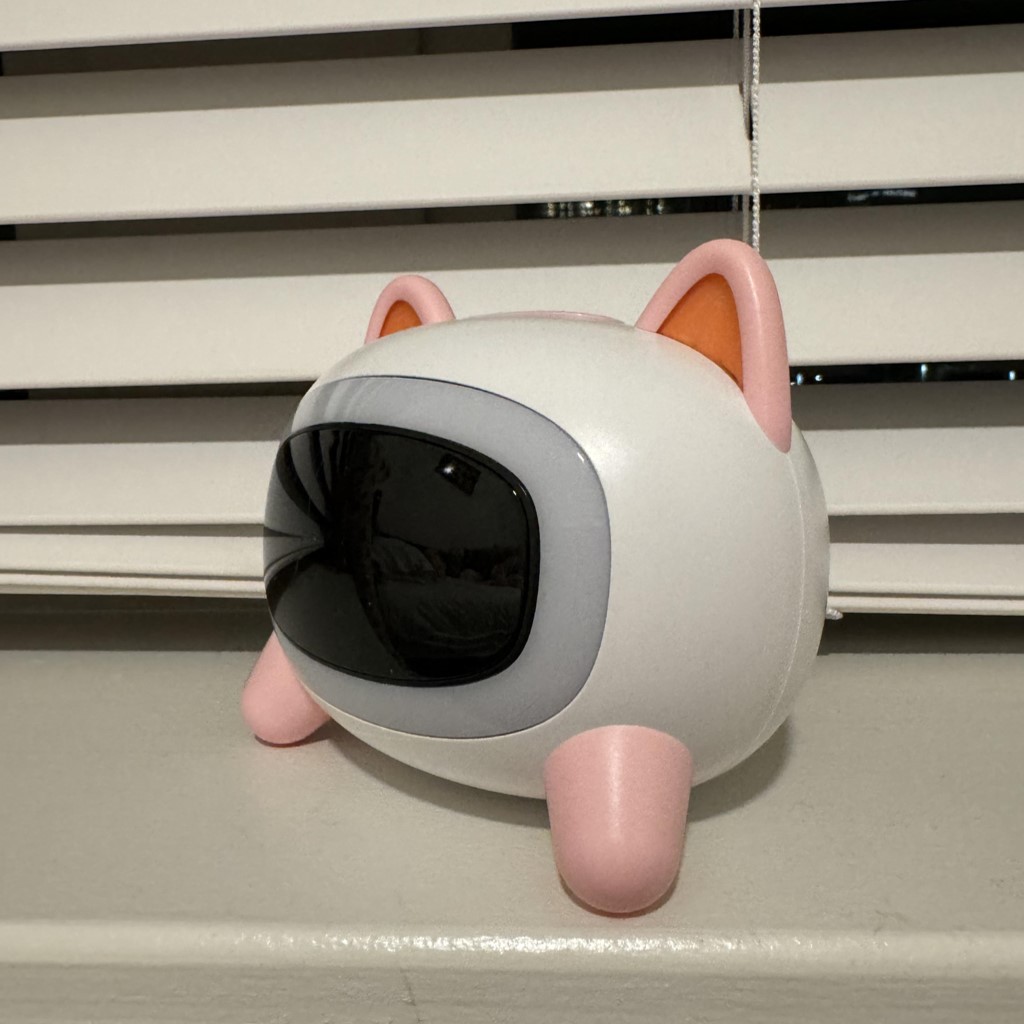} 
    \includegraphics[width=0.12\linewidth,height=0.12\linewidth,trim={0cm 0cm 0cm 0cm},clip]{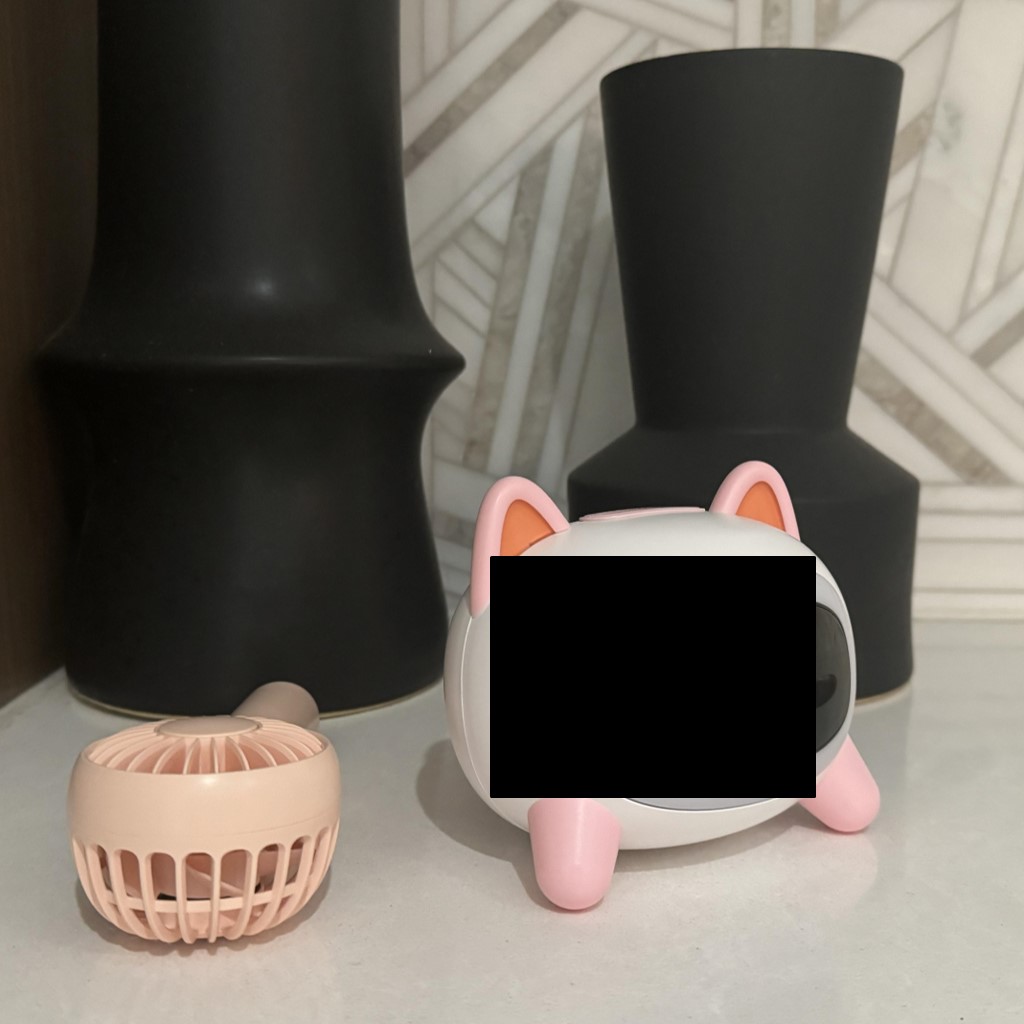}
    \includegraphics[width=0.12\linewidth,height=0.12\linewidth,trim={0cm 0cm 0cm 0cm},clip]{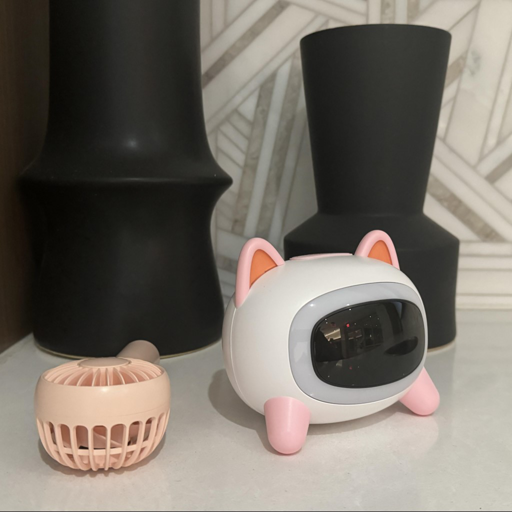} 
    \hspace*{0.1em} \vline  \hspace*{0.4em}
    \includegraphics[width=0.12\linewidth,height=0.12\linewidth,trim={0cm 0cm 0cm 0cm},clip]{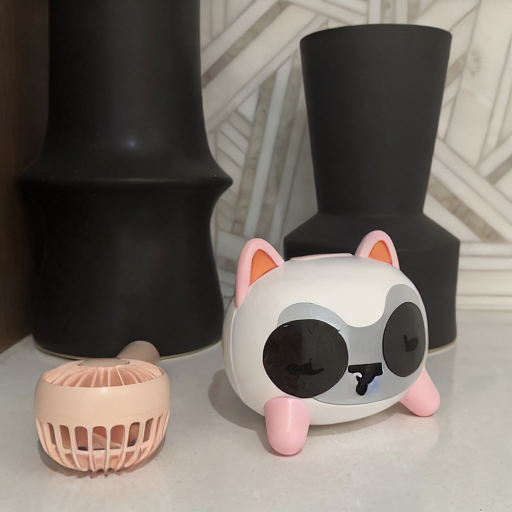}
    \includegraphics[width=0.12\linewidth,height=0.12\linewidth,trim={0cm 0cm 0cm 0cm},clip]{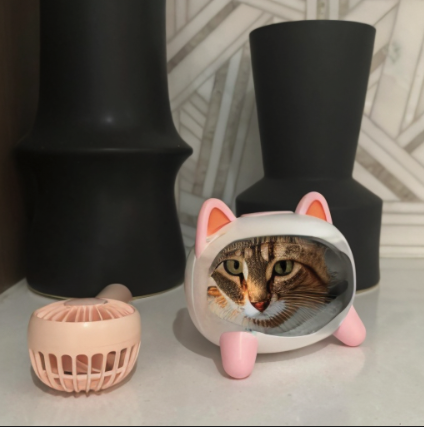}
    \includegraphics[width=0.12\linewidth,height=0.12\linewidth,trim={0cm 0cm 0cm 0cm},clip]{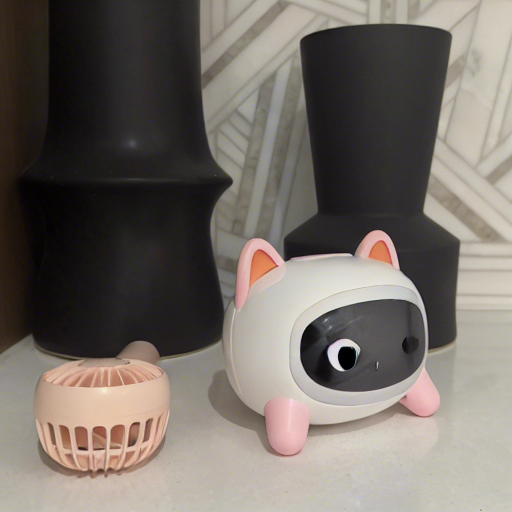}
    \includegraphics[width=0.12\linewidth,height=0.12\linewidth,trim={0cm 0cm 0cm 0cm},clip]{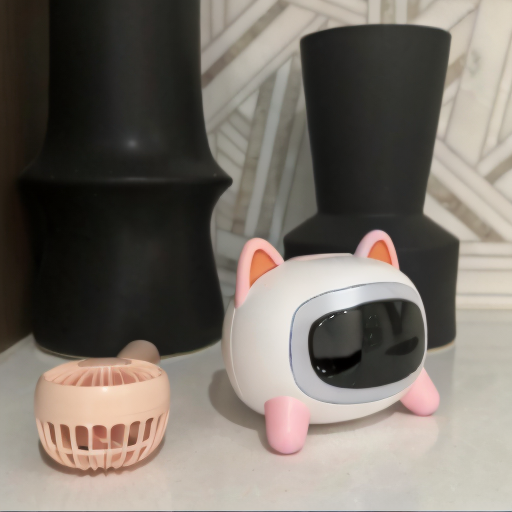} 
    
\caption{Inpainting results of different missing regions using FaithFill compared to some state-of-the-art techniques. While state-of-the-art techniques provide high-quality plausible inpainting results, they may not be faithful to the object. This is observed in both methods that do not use a reference image (\textit{Stable Inpainting} and \textit{Blended Latent Diffusion}), and methods that use a single reference/exemplar image (\textit{Paint By Example}). The first row is a sample image from the DreamBooth dataset, and the second row is a sample image from our proposed FaithFill dataset. More qualitative examples for additional state-of-the-art techniques are presented later in the paper.}
\label{fig:Intro}
\end{figure*}
\section{Introduction}
\label{sec:intro}


The success of generative diffusion models has paved the way for successful image editing, including image inpainting. Image inpainting is the go-to solution for recovering occluded or corrupted image/object regions.
Using text-to-image models such as Stable Diffusion \cite{Rombach_2022_CVPR_stable_diffusion} works well for inpainting, but while it produces a plausible and realistic result, that result may not preserve shape, color, or texture features of the original foreground or background. This is due to lack of contextual information while inpainting. We preserve contextual information by finetuning on a reference image. 





Typically, generative models - after some finetuning iterations - will give an impressive result, but will not be faithful to neither the foreground or background. We define the preservation of shape, texture, and color as faithfulness. The first technique that introduced faithfulness (`authenticity') in generative inpainting models is RealFill~\cite{tang2023realfill}. RealFill requires multiple reference images that have very similar viewpoints to the target viewpoint to achieve such authenticity. 

Most finetuning is based on pluralistic approaches, \ie~using multiple reference images, or pretraining on a similar domain dataset. The former is more computationally efficient. While RealFill is an example of the former, Paint-By-Example~\cite{yang2022paint} is an example of the latter. Paint-By-Example fuses an object from a reference image to the target image in a realistic manner using a diffusion-based framework. Paint-By-Example conditions the model on the reference image at inference time, but finetunes on the entire OpenImages~\cite{openimages} dataset at training time. 
While Paint-By-Example uses a reference image after finetuning, SmartBrush~\cite{Xie_2023_CVPR_Smart_Brush} uses the target object mask and text conditioning to inpaint in an authentic way. Using such a target object mask preserves the background. 

The challenge arises when we have a \textit{single} reference image for finetuning (most computationally efficient scenario) due to less prior knowledge compared to the pluralistic approaches. The correspondence between the reference and target images might vary quite extensively due to different backgrounds, viewpoints, object poses, object shapes, and lighting conditions. One-shot finetuning is prone to underfitting \wrt the reference image, causing distortion of the object's shape, color, texture, or not preserving the background during the reconstruction of the masked region.  

In this work, we propose FaithFill, a generative in-painting technique that is faithful to both foreground and background objects in the image (Figure~\ref{fig:Intro}), \textit{and} only requires a single reference image in order to do so. FaithFill finetunes on a single reference image, and requires a minimal text prompt \vs~long textual descriptions or prompt tuning. We overcome the risk of one-shot finetuning by generating multiple object views using Neural Radiance Field (NeRF) based models~\cite{liu2023syncdreamer,liu2023one}. This gives FaithFill additional flexibility in view point change between the reference and the to-be-inpainted (target) image compared to RealFill~\cite{tang2023realfill}. Concurrent work LeftRefill~\cite{cao2024leftrefill} produces multiple views in an autoregressive manner. In contrast, we generate the views in one-shot using NeRFs. 




We demonstrate more faithful inpainting results compared to state-of-the-art, and concurrent work of LeftRefill\cite{cao2024leftrefill} for most instances both quantitatively and qualitatively. Evaluation is based on the quality of image generation \wrt~ that of the ground truth image using (a) standard similarity metrics used in the recent inpainting literature, (b) human judgement, and (c) GPT evaluation. 

We summarize our contributions as follows.

\begin{itemize}
    \item We propose FaithFill, a finetuning pipeline that is able to faithfully inpaint objects using a single reference image. Faithfulness is defined to be reservation of shape (pose can change for deformable objects), color, and texture.
    \item We propose the FaithFill Dataset, a dataset of image pairs of 45 objects taken under different lighting conditions, from different viewing angles, and with different background settings. Each pair consists of a reference image and a target image.
    \item FaithFill demonstrates superior performance to \sota~on the six standard image similarity metrics, human judgement, and GPT evaluation.
\end{itemize}

\section{Related Works}
\label{sec:related work} 

\paragraph{Diffusion Models.} 
Image generation has been revolutionized with the introduction of diffusion models. There are a number of works for image-to-image generation such as DDPM \cite{NEURIPS2020_ddpm}, DDIM \cite{song2021denoising_ddim}, and text-to-image generation such as DALL-E \cite{ramesh2021zeroshot}, Imagen \cite{NEURIPS2022_ec795aea_Imagen}, Stable Diffusion \cite{Rombach_2022_CVPR_stable_diffusion}. The main breakthrough introduced by diffusion models lies in reversing the Markov denoising process. The models based on this principal achieve \sota~results in many computer vision tasks. Text-to-Image diffusion models have been trained on the LAION-5B \cite{schuhmann2022laion5b} dataset that enables models to use this as prior information for further finetuning. Finetuning these models leads to impressive results for image editing, controllable image generation, and image personalisation~\cite{Ruiz_2023_CVPR_DreamBooth}, video generation, pose generation \cite{bartal2024lumiere, ho2022video}, texture generation \cite{chen2023text2tex}, panoramas \cite{wu2024panodiffusion}, 3D meshes \cite{liu2023one2345, liu2023syncdreamer} \etc

\paragraph{Generative Diffusion Based Models for Image Inpainting.} 
Image-inpainting is an important problem as filling up the missing regions within an image has plethora of uses. The inpainting problem has been long studied  \cite{ImageInpainting, PatchMatch} dating back prior to deep learning based models. In deep learning based models, a neural network is trained to complete the missing regions. Furthermore, generative models leveraged image prior(s) for completion of missing region(s). More recently, text-to-image models use text priors to fill in missing image regions.  

RePaint~\cite{Lugmayr_2022_CVPR} is one of the early works inspired to use the fundamental principal of diffusion models by iteratively denoising the Gaussian noise to fill in the missing regions for any given shape of the target mask. This model is based on the image-to-image diffusion model, DDPM \cite{NEURIPS2020_ddpm} that conditions on a given image region. Stable Inpainting~\cite{Rombach_2022_CVPR_stable_diffusion} uses stable diffusion to generate samples from a noisy latent distribution. Stable Inpainting requires additional conditioning on the target masks in addition to that of text. Blended Diffusion~\cite{Avrahami_2022_CVPR} leveraged the combination of CLIP \cite{Radford2021LearningTV_CLIP} and Denoisining Diffusion Probabilistic Model (DDPM) \cite{NEURIPS2020_ddpm} for prompt-based image editing. Combining the pretrained text-image model such as CLIP with DDPM enabled mitigation against the adversarial examples as they blend the text latent along with the image at each denoising step. This method uses a multistep blending process to fill the masked regions. Later, the authors also extended this work using a Text-to-Image model instead of the DDPM \cite{avrahami2023blendedlatent}. GLIDE~\cite{pmlr-v162-nichol22a} uses the mask conditioning in addition to the text conditioning for the model to mask the image regions. It uses an ablated diffusion model \cite{dhariwal2021diffusion} along with a transformer \cite{NIPS2017_transformer} in addition to an upsampler to obtain higher resolution images. 

The closest works to our proposed FaithFill are Paint-By-Example \cite{yang2022paint}, RealFill \cite{tang2023realfill}, and very recently LeftRefill\cite{cao2024leftrefill}. Paint-By-Example is similar because it targets blending the target image with an object from the reference image. However, Paint-By-Example uses pretraining the augmentations on an entire dataset while we only finetune on a single reference image. RealFill or DreamBooth-Inpaint \cite{Ruiz_2023_CVPR_DreamBooth,tang2023realfill} use 3-5 reference images for the image completion making it either requiring $n>3$ reference images, or failing on complex scenarios. In contrast, we finetune on a single reference image. LeftRefill uses Novel View Synthesis (NVS) from a single reference image in an autoregressive manner to generate the right view from a left view for the purpose of image stiching. In contrast, we do not require iterative NVS, we generate the different views in one-shot using NeRFs and address objects with different backgrounds, lighting, and larger viewpoint change. 

\paragraph{Generative Adversarial Models for Image Inpainting.} Prior to the advent of diffusion models, GANs have been employed to tackle the image inpainting task. One of the early methods that did this is Context Encoders (CE) \cite{Pathak_2016_CVPR} that attempts to perform image inpainting using semantically consistent content. Nevertheless, this method is constraint to filling only square holes located in the center of an image. GAN inversion inpainting methods have been employed in \cite{Yeh_2017_CVPR, Lahiri_2020_CVPR, Wang_2022_CVPR, Yildirim_2023_ICCV}, where the latent space of a pretrained GAN is searched for a latent code representing the masked image. Then, the image is reconstructed by reversing the latent code to get the inpainted image. Other deep learning methods have incorporated patch borrowing and patch replacement operations \cite{Yang_2017_CVPR, Yu_2018_CVPR, Zeng_2019_CVPR, Xiong_2019_CVPR, Liu_2019_ICCV, Zeng_2021_ICCV}. PEPSI \cite{Sagong_2019_CVPR} adopts a single decoder-encoder network for semantic inpainting, in contrast to the use of cascading fine-to-coarse networks. Li \etal~\cite{Li_2019_ICCV} introduced a visual structure reconstruction layer to entangle the generation of the image structure and the corresponding visual content in a progressive fashion. \cite{Yi_2020_CVPR} proposes a model for inpainting ultra high-resolution images up to 8K. \cite{Zhao_2020_CVPR} introduces a conditional image-to-image generation framework for multiple-output (diverse) image inpainting. Other methods focused on improving texture and structure via dual path inpainting \cite{Liao_2021_CVPR, Guo_2021_ICCV, Wang_2022_CVPR}. Jain \etal~ \cite{Jain_2023_WACV} combine a coarse-to-fine GAN-based generator with fast Fourier convolution layers to improve both structure and texture generation. Sragsyan \etal~ introduced MI-GAN \cite{Sargsyan_2023_ICCV}, an image inpainting model for mobile devices. \cite{Jam_2021_WACV} Integrates Wasserstein Generative Adversarial Network (WGAN) with a reverse masking operator to focus only on the masked part and maintain the rest of the image unchanged. \cite{Liu_2021_CVPR} proposed SPDNorm to alter the random noise vector of the vanilla GAN network in order to take context constraints into consideration. Ni \etal \cite{Ni_2023_CVPR} proposed a language-guided Image inpainting model based on the DF-VQGAN.

\begin{figure*}[t]
    \centering
    \includegraphics[width=1.0\linewidth]{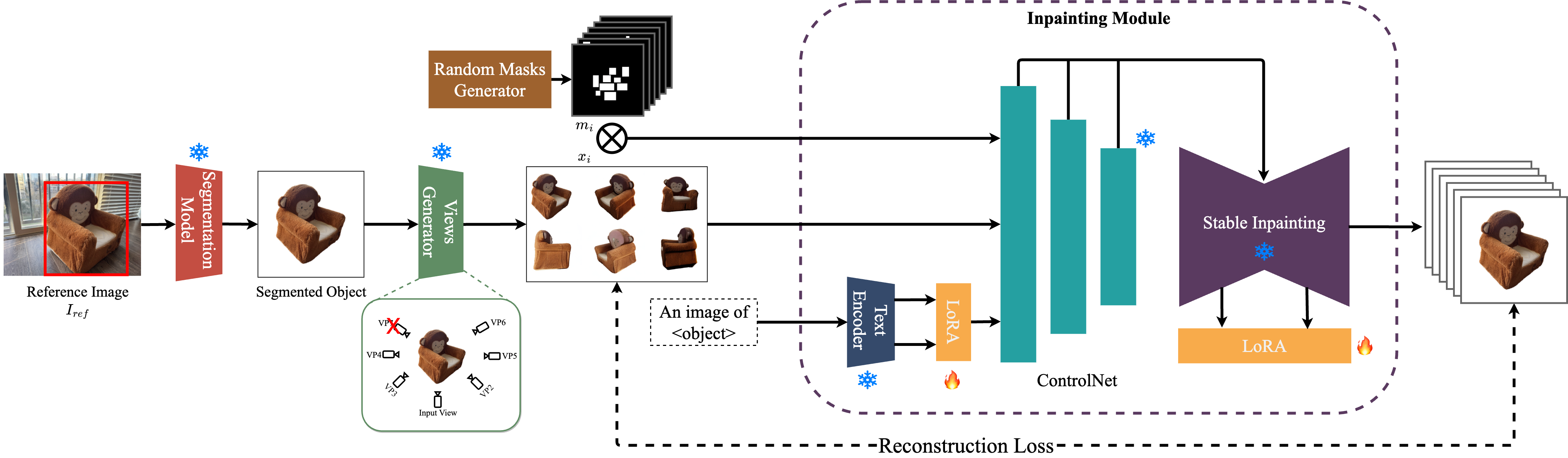}
    \caption{\textbf{FaithFill Finetuning Pipeline.} This figure presents the schematic overview of our finetuning pipeline. Given an input image $I_{ref}$ we generate $n$ different images \{$x_{1}, x_{2}....x_{n}$\} from different viewpoints (VP) using a view generator based on NeRFs. The views $\{ x_{1}, x_{2}....x_{n} \}$ are then multiplied with randomly generated masks $\{ m_{1}, m_{2}....m_{n} \}$. The randomly masked views are used as an input along with the text to the Inpainting Module. In this module we finetune the LoRA adapted layers instead of finetuning the whole model. Finetuning is governed by a reconstruction loss with respect to the unmasked generated views.}
    \label{fig:pipeline}
\end{figure*}

\section{Background: Diffusion Models}
\label{sec:background}

Diffusion Models are generative models that gradually invert the Markovian forward process for denoising a Gaussian distribution to generate an image. Latent diffusion models \cite{Rombach_2022_CVPR_stable_diffusion} work on a latent representation to reduce the computation cost. In this method text prompts are encoded using a pretrained CLIP model. An input image $x_{0} \in \mathbb{R}^{H \times W \times 3}$ is converted into the latent space as $z_{0}$ using a pretrained autoencoder $\varepsilon$ (\ie, $z = \varepsilon(x)$). The input latent representation goes through the Markov forward process of adding noise for $t$ time steps to produce $z_{t}$. The noisy latent representation $z_{t}$ is obtained as presented in Equation~\ref{eq:noisy_latent}, where the noise $\epsilon \sim \mathcal{N}(\textbf{0}, \mathbf{I})$ and ${\{\alpha_{t}\}_{t=1}^{T}}$ and $\alpha_{t}$ is the noise scheduler.

\begin{equation}\label{eq:noisy_latent}
    z_{t} = \sqrt{\alpha_{t}}z_{0} + (\sqrt{1-\alpha_{t}})\epsilon
\end{equation}

The overall learning objective given the encoded text prompt $\tau(p_{t})$ for a text-to-image model is presented in Equation~\ref{eq:simple_diffusion}.

\begin{equation}\label{eq:simple_diffusion}
    \mathcal{L} =  \mathbb{E}_{z_{0}, t, \tau(p), \epsilon \sim \mathcal{N}(0, 1)} \Big[ \lVert \epsilon - \epsilon_{\theta}(z_{t}, t, \tau(p)) \rVert^{2}_{2} \Big]
\end{equation}

For the inference, inversion starts from $z_{t} \sim \mathcal{N}(\textbf{0}, \mathbf{I})$ and reduces noise at $t-1$ timestep, ($z_{t-1}$) using the denoiser $\epsilon_{\theta}$ finally obtaining the target denoised $z_{0}$ iteratively for $t$ time steps. While any denoiser could be used here, we use the standard U-Net as the denoiser. The $z_{0}$ is the latent representation of the target image which is finally decoded using the decoder $\mathcal{D}$ to the image space.


\section{Method: FaithFill}
\label{sec:methods}

In this work we are proposing a finetuning framework for a reference-based image inpainting technique for a target image to make it plausible, photo-realistic, and faithful to object attributes in the reference image. In this section we discuss FaithFill's finetuning and inference pipelines. The input to the finetuning pipeline is the reference image ($I_{ref} \in \mathbb{R}^{H \times W \times 3}$), and the input to the inference pipeline is the masked target image ($I_{tgt} \in \mathbb{R}^{H \times W \times 3}$). Our FaithFill finetuning pipeline is depicted in Figure~\ref{fig:pipeline}. This includes a segmentation module to extract the object from the reference image, a diffusion based view generation module to generate multiple views of the extracted object, followed by an inpainting module that reconstructs the masked image. This section will describe the different modules of this pipeline.

\subsection{FaithFill Finetuning}

\paragraph{Segmentation Module.} 
This module receives as input the reference image  $I_{ref} \in \mathbb{R}^{H \times W \times 3}$. It first determines the object of interest, \ie~the object that needs to be inpainted using $I_{tgt}$ \ie, $\{ I_{ref} \cap I_{tgt} \}$. We then use the Segment Anything Model (SAM) \cite{kirillov2023segany} to extract the object. The segmented/extracted object is presented to a diffusion based multiple view synthesis module. The background is removed from different views to preserve the homogeneity and natural blending in the target image $I_{tgt}$ when inference happens. 

\paragraph{View Generation Module.}
We perform multi-view synthesis of the extracted object from the Segmentation module using diffusion based NeRFs, this is inspired by Liu \etal~\cite{liu2023one}. Liu \etal~\cite{liu2023one} use the Zero123 \cite{Liu_2023_ICCV} model, a viewpoint conditioned 2D diffusion model for the generation of multi-view images from the input that transforms these views to a 3D space. This is achieved by finetuning a stable diffusion model. Our aim is is not to create a 3D mesh from a single image but rather get multiple 2D views for finetuning later modules in our pipeline. We do not train or finetune the NeRFs model, rather we use its inference as a view generation module to generate $N=6$ ($5~plus~original$) views as presented in Figure~\ref{fig:pipeline}. Let the set of images with different viewpoints (VP) be denoted as $x \in \mathcal{X}_{VP}$, where $\mathcal{X}_{VP} = \{x\}_{n=1}^{N}$. Using this module enables us to use reference images that have more flexible viewpoints than state-of-the-art reference based inpainting methods~\cite{tang2023realfill, yang2022paint}. This module could be switched out for any multi-view synthesis module, \eg~\cite{liu2023syncdreamer}.

Once the views are generated, a Random Mask Generator creates one mask per view $\{m\}_{n=1}^{N}$. Each mask is randomly centered and masks at least $ratio$ percentage of the view image. The mask consists of a random number of rectangles, each having a random width between $0$ and $w*ratio$, and a random height between $0$ and $h*ratio$ such that the masking ratio is achieved.

\paragraph{Inpainting Module.} 
The inpainting module consists of a CLIP text encoder~\cite{Radford2021LearningTV_CLIP}, a ControlNet~\cite{Zhang_2023_ICCV} adapter, and stable inpainting pipeline that uses a U-Net denoiser. The inpainting module takes as input the generated views alongside corresponding randomly generated masks and textual description. 

The multiple viewpoint images $x_n$ are masked with an inverted mask $m_n$ using a Hadamard product \ie, it's computed as $x_n \odot (1-m_n)$. The input to the ControlNet pipeline are $\{x_n \odot (1-m_n), m_n, \tau(p) \}$, where $\tau(c)$ is the text embedding, and $\tau(.)$ is CLIP text encoder for the prompt $p$. The output of the ControlNet is then passed to the Stable Inpainting pipeline for filling in the missing region. The ControlNet adapter provides additional control to resist updates to the unmasked regions of the different views. 

We use a Low Rank Adaptation Technique (LoRA)~\cite{hu2022lora} for finetuning the U-Net and the CLIP text encoder in the inpainting module, instead of finetuning of the complete module that is computationally expensive. LoRA injects the trainable low rank residual matrices in addition to the network weight matrices. The pretrained weight matrix $W \in \mathbb{R}^{n \times n}$ is then updated with the low rank decomposition matrices as $W + \Delta W = W + BA$, where $B \in \mathbb{R}^{n \times r}$ and $A \in \mathbb{R}^{r \times n}$, $r<<n$. During the finetuning process, the low rank matrices $A$ and $B$ are updated while the network weights $W$ are frozen. The loss function that governs the finetuning is presented in Equation~\ref{eq:faithfill_diffusion}.

\begin{equation}\label{eq:faithfill_diffusion}
    \mathcal{L} =  \mathbb{E}_{x, t, \tau(p), \epsilon \sim \mathcal{N}(0, 1)} \Big[ \lVert \epsilon - \epsilon_{\theta}((x \odot (1-m)), m, t, \tau(p)) \rVert^{2}_{2} \Big]
\end{equation}

\subsection{FaithFill Inference}

\paragraph{Inpainting Module.} 
At inference time, the goal is to complete the missing regions of the ${I_{tgt}}$. The input to the pipeline is $\{ I_{tgt}, m, \tau(p) \}$, where $p$ is the same text prompt used for finetuning the reference image \ie, \textit{``an image of the <object class>}.'' The inpainting module with the modified weights based on the finetuning on the reference image is used to inpaint the masked target image. This inference module is Stable-Inpainting based on Stable-Diffusion v2 pipeline with a DPMS sampler \cite{lu2022dpm}.

In an ideal scenario, the reconstructed image must preserve the regions other than the regions that need to be inpainted, therefore we restrict alterations to the missing regions by employing a binary mask $\{0,1\} \in \mathbb{R}^{H \times W}$. Following \cite{Lugmayr_2022_CVPR} and other inpainting works, $0$ denotes the regions to skip whereas $1$ denotes the regions to fill. 

\section{Experiments}

In this section we first introduce the benchmark datasets used for our experiments, followed by the evaluation metrics used to compare inpainted results against ground-truth. We then present the implementation details for reproducing our experimental setup and results. Finally, we present and discuss qualitative and quantitative inpainting results of similarity metrics, human judgement, and GPT judgement for FaithFill and comparison to previous state-of-the-art. 

\subsection{Benchmark Datasets}
\label{subsec:FaithFill Dataset}
We evaluate on the DreamBooth dataset \cite{Ruiz_2023_CVPR_DreamBooth} that contains multiple reference images for a single subject/object (total of 30 subjects/objects). Each subject/object has $3-5$ casually taken reference images. 
We randomly sample a pair of images from each subject/object where we randomly assign reference and target image roles. This dataset presents multiple views, backgrounds, lighting settings \etc 

In addition to the DreamBooth dataset, we created our FaithFill Dataset for a more comprehensive evaluation on a larger number of images with larger viewpoint variations. The FaithFill dataset consists of 45 objects where they were selected for their intricate structures and varied textures. For each object, we captured two images depicting it from different viewpoints, under different lighting, and against different backgrounds. Each pair constitutes a reference image and a target image. This dataset will be made publicly available.

\subsection{Implementation Details}
 
The multi-view generation module is used in the inference time with the settings mentioned by the authors in \cite{liu2023one2345}. Further, the number of iterations to finetune our model depends on the dataset used. We finetune 1100 iterations or timesteps for the DreamBooth dataset on a single 40GB NVIDIA A100 GPU. We use 1500 iterations or timesteps for the FaithFill dataset. We hypothesize that this is because of a larger viewpoint difference between reference and target images in the FaithFill dataset. 

We set the masking percentage of the Random Mask Generation module to 50\%. The text-prompt used as input to the text encoder is of the form `An image of \textit{<object>}'.

As mentioned in Section~\ref{sec:methods}, we use LoRA based models. The LoRA rank for both the datasets is set to be 4. The guidance scale for inference is set to be 7.5. The learning rates are set to be 5e-4. For the baseline experiments the hyperparameters are kept as is recommended by the respective authors.

\subsection{Evaluation}

We evaluate on both the publicly available DreamBooth and our proposed FaithFill datasets qualitatively and quantitatively. We compare against state-of-the-art using image similarity metrics, human judgement, and GPT evaluation. For image similarity metrics, we use low-level perceptual similarity that computes the texture and color, mid-level semantic differences such as layout and pose, as well as high level differences that engrave more high-level attributes. The low-level SSIM is a patch based similarity matrix that computes the difference in structural similarity. SSIM fails to capture some human perception nuances, thus LPIPS was introduced. LPIPS computes the feature distance between the two patches. PSNR is another low-level pixel-wise image similarity comparing the signal to that of the background noise. For mid-level semantic differences, the images are evaluated on DreamSIM, that aims to have evaluation standards similar to that of human perception. The use of the DreamSIM enables to bridge the gap between the low-level and high level semantics. DINO and CLIP are used for high-level semantic differences. CLIP differentiates between semantic consistency while DINO is used for the semantic parts. CLIP computes the mean cosine similarity of the embeddings computed using CLIP between the ground truth and the generated image, while \cite{Ruiz_2023_CVPR_DreamBooth} introduced DINO where it computes the mean cosine similarity between the ViT-S/16 features between the generated and ground truth image.  

\begin{figure*}[t!]
    \centering
    \begin{subfigure}
        \centering
        \includegraphics[width=0.48\linewidth, height=0.34\linewidth]{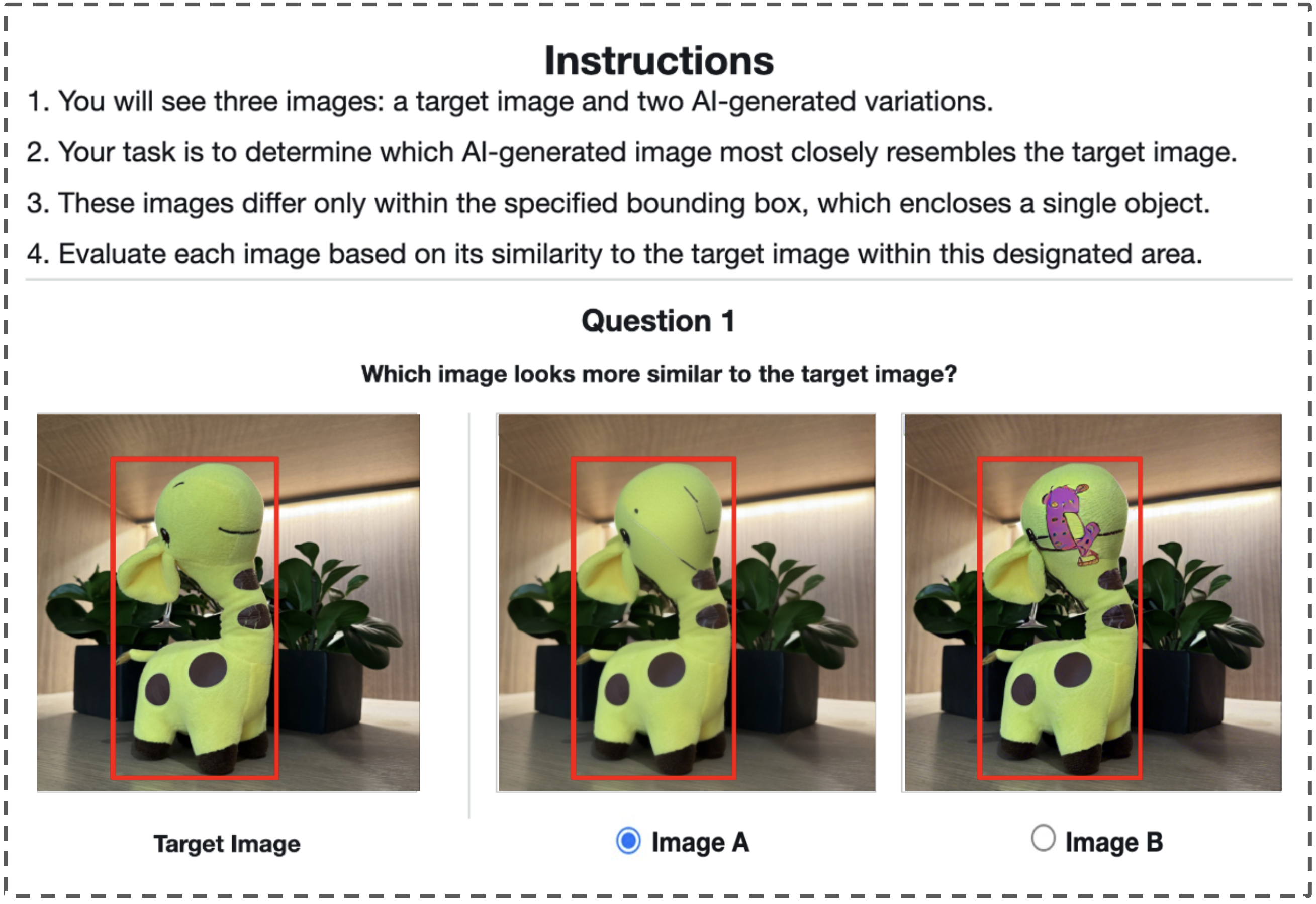}
    \end{subfigure}%
    \begin{subfigure}
        \centering
        \includegraphics[width=0.48\linewidth,height=0.34\linewidth]{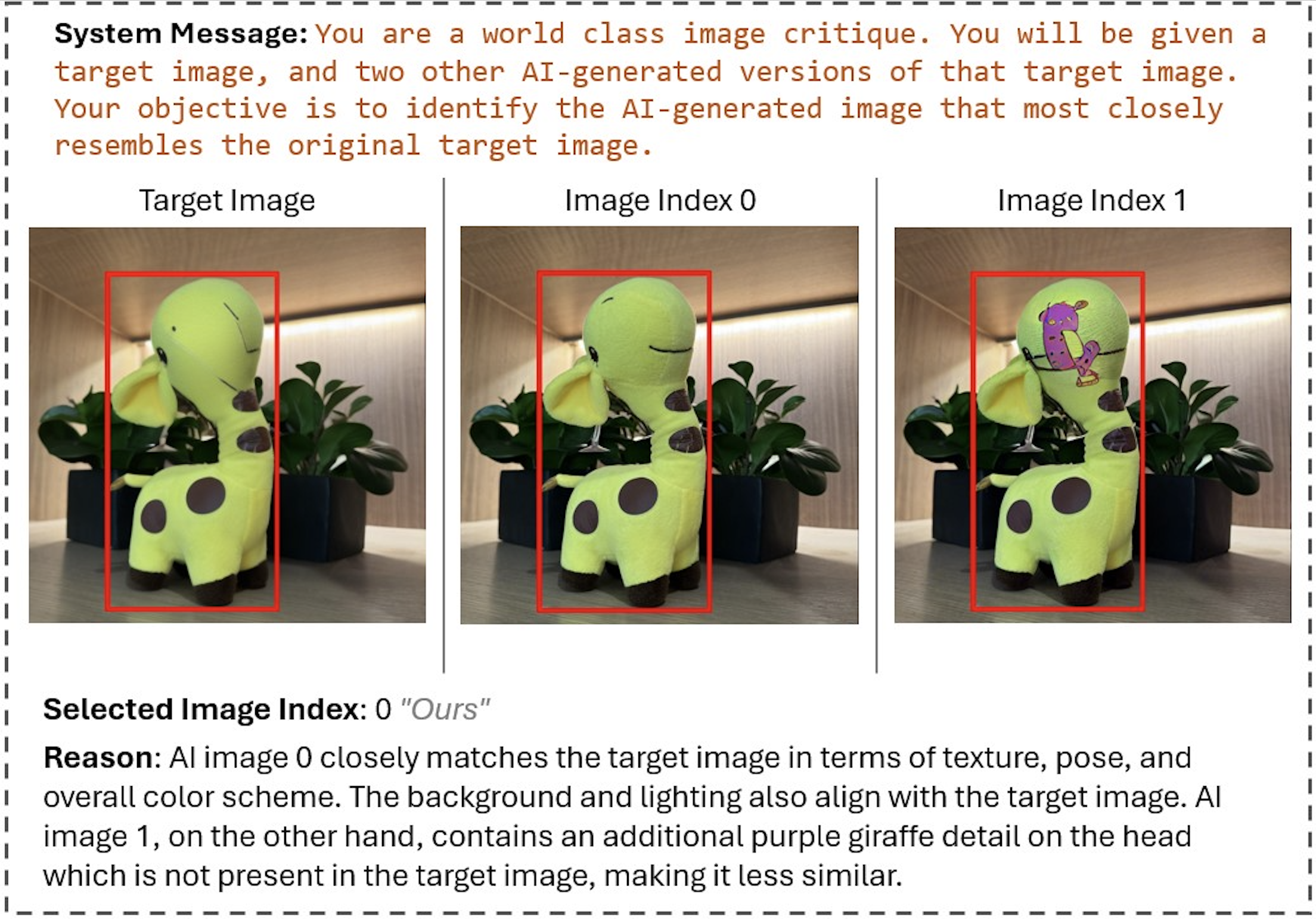}
    \end{subfigure}%
    \hfill
    \caption{\textbf{AMT Interface (Left).} This is a screenshot from the Amazon Mechanical Turk interface that we used to launch our user study. \textbf{GPT-4o Setup (Right).} We asked GPT-4o to compare two AI generated images (ours vs a baseline-generated image) and decide which one is more similar to the target image. It was prompted to provide the index of the selected image as well as the reason for the selection. This figure presents a sample result.}
    \label{fig:gpt_reasoning}
\end{figure*}

In addition to the presented similarity metrics and quantitative results, we conducted a user study and a GPT evaluation comparing each FaithFill generation to each state-of-the-art method generation \wrt~the ground-truth target image. For the user study, we use the Amazon Mechanical Turk (AMT) crowdsourcing marketplace to recruit crowd workers. We accept AMT workers who had previously completed at least 1000 tasks (a.k.a HITs), and maintained an approval rating of at least 98\%. We compensate the work of all crowd workers who participated in our tasks. In the user study, we ask a random set of nine evaluators for each comparison to determine which image most closely resembles the target image. Each subtask presents the worker with one target image and two generated images, one is FaithFill generated, and the second is a competing state-of-the-art method generation. A sample interface question is presented in Figure~\ref{fig:gpt_reasoning}. The worker is asked to select which of the two images is more similar to the target image. We post all HITs simultaneously,  while randomizing the presentation order of the FaithFill \vs other images. We allot a maximum of ten minutes to complete each HIT and paid \$0.10 per HIT. For the GPT evaluation, we presented GPT-4o with the same question setup we described for the user study. We run the evaluation three times on each pair of images. Figure~\ref{fig:gpt_reasoning} shows a sample output from the GPT-4o study.

\begin{figure*}[!t]
    \centering 
    \setlength{\arrayrulewidth}{1.5pt}

    \rotatebox[origin=c]{0}{
    \begin{tabular}{cccccccc}
      \hspace*{1.5em}
      \parbox{1.5cm}{\centering \textit{Reference} \\ \textit{Image}} &
      \parbox{1.5cm}{\centering \textit{Masked} \\ \textit{Target}} &
      \parbox{1.5cm}{\centering \textit{Target} \\ \textit{Image}} &
      \hspace*{-0.2em}
      \parbox{1.5cm}{\centering \textit{Stable} \\ \textit{Inpainting FT}} &
      \parbox{1.5cm}{\centering \textit{LeftRefill}} &
      \hspace*{-1.2em}
      \parbox{1.5cm}{\centering \textit{Paint By} \\ \textit{Example}} &
      \parbox{1.5cm}{\centering \textit{\textbf{FaithFill}} \\ \textit{\textbf{(Ours)}}}
    \end{tabular}
  }

    {\rotatebox[origin=c]{90}{\parbox{1cm}{\centering\textit{\textcolor{white}{xxx}} \\ \textit{Dog}}\hspace*{-4em}}} 
    \includegraphics[width=0.12\linewidth,height=0.12\linewidth,trim={0cm 0cm 0cm 0cm},clip]{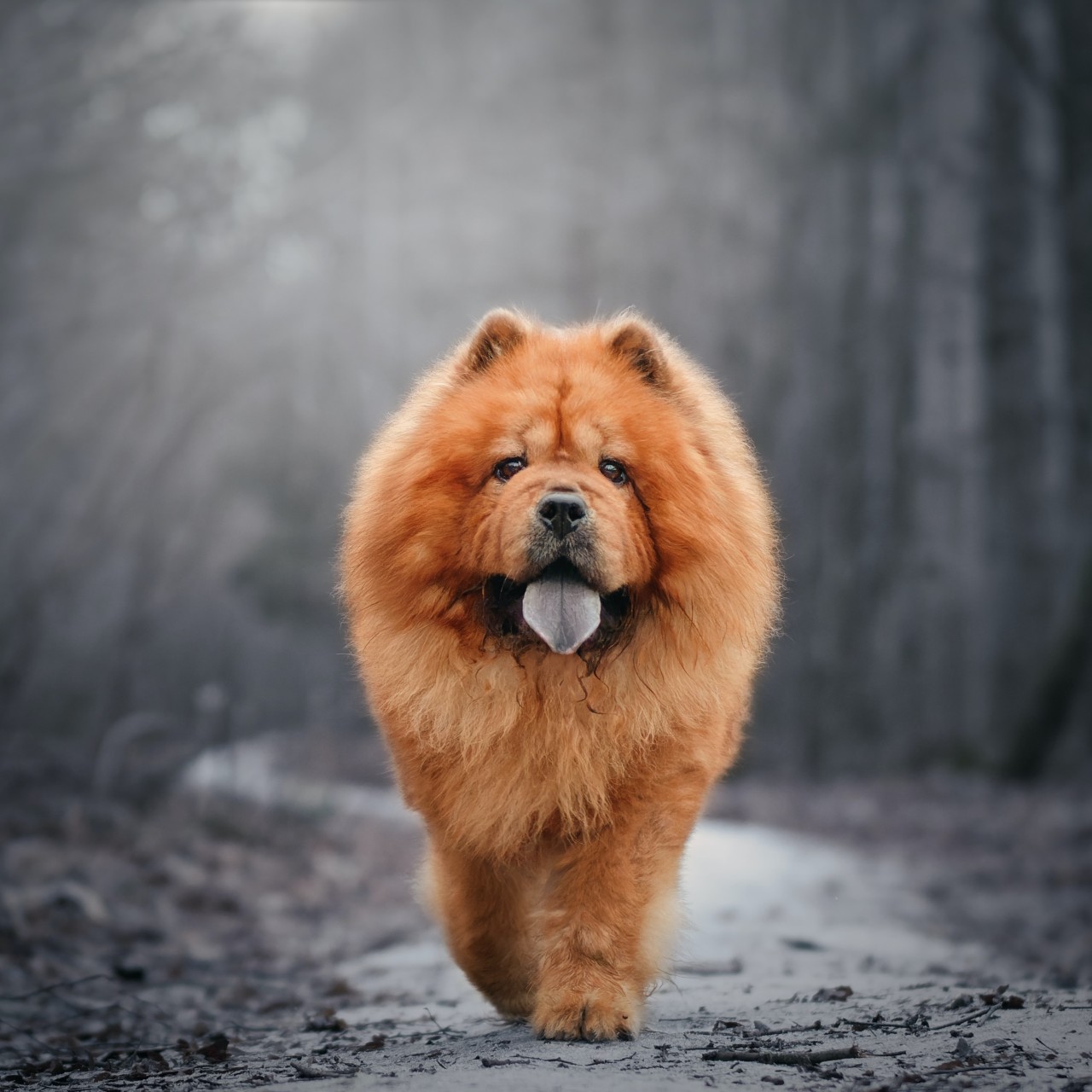} 
    \includegraphics[width=0.12\linewidth,height=0.12\linewidth,trim={0cm 0cm 0cm 0cm},clip]{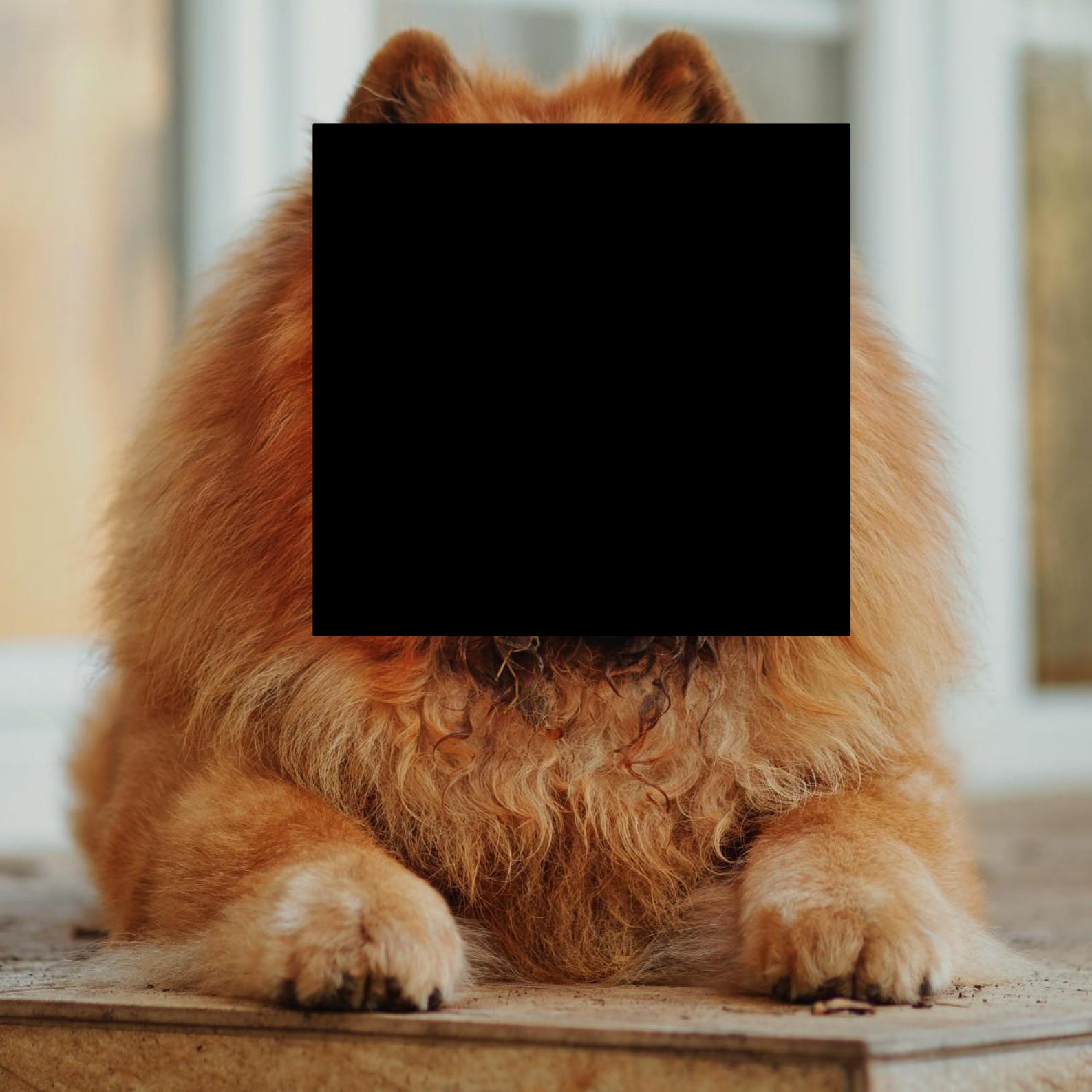} 
    \includegraphics[width=0.12\linewidth,height=0.12\linewidth,trim={0cm 0cm 0cm 0cm},clip]{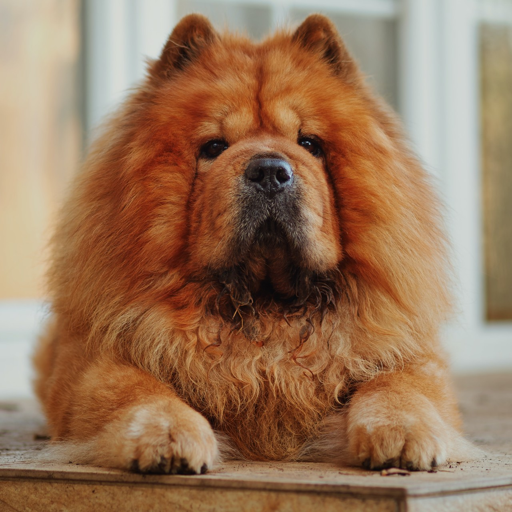} 
    \hspace*{0.1em} \vline  \hspace*{0.4em}
    \includegraphics[width=0.12\linewidth,height=0.12\linewidth,trim={0cm 0cm 0cm 0cm},clip]{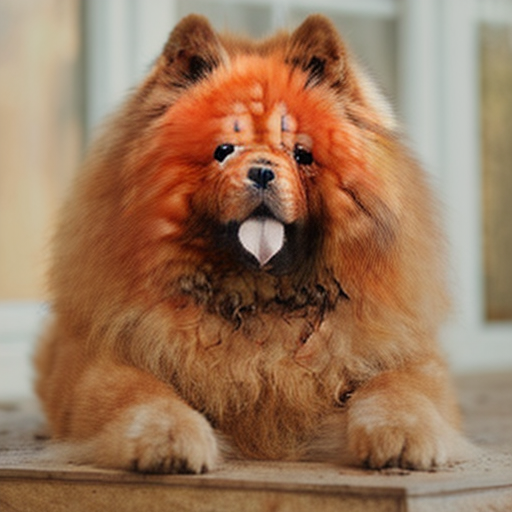}
    \includegraphics[width=0.12\linewidth,height=0.12\linewidth,trim={0cm 0cm 0cm 0cm},clip]{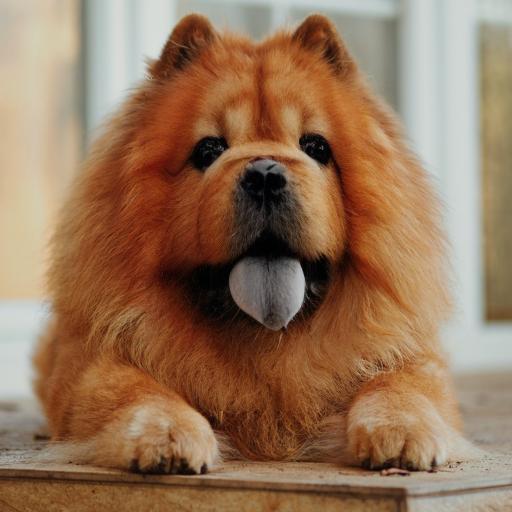} 
    \includegraphics[width=0.12\linewidth,height=0.12\linewidth,trim={0cm 0cm 0cm 0cm},clip]{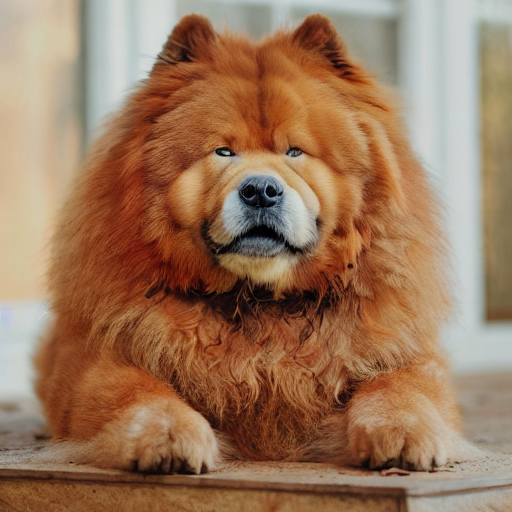}
    \includegraphics[width=0.12\linewidth,height=0.12\linewidth,trim={0cm 0cm 0cm 0cm},clip]{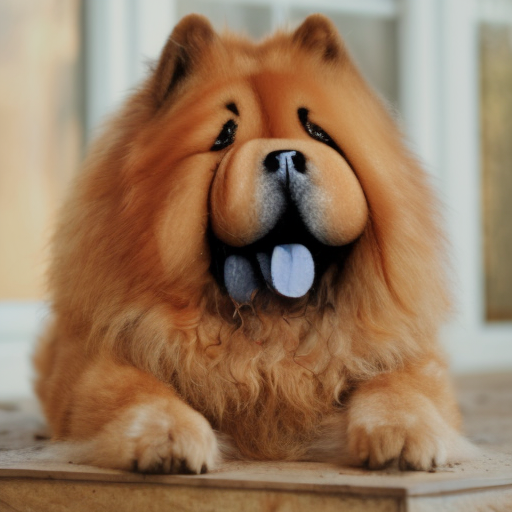} \\[0.2em]
    
    {\rotatebox[origin=c]{90}{\parbox{1cm}{\centering\textit{\textcolor{white}{xxx}} \\ \textit{Bag}}\hspace*{-4em}}}  
    \includegraphics[width=0.12\linewidth,height=0.12\linewidth,trim={0cm 0cm 0cm 0cm},clip]{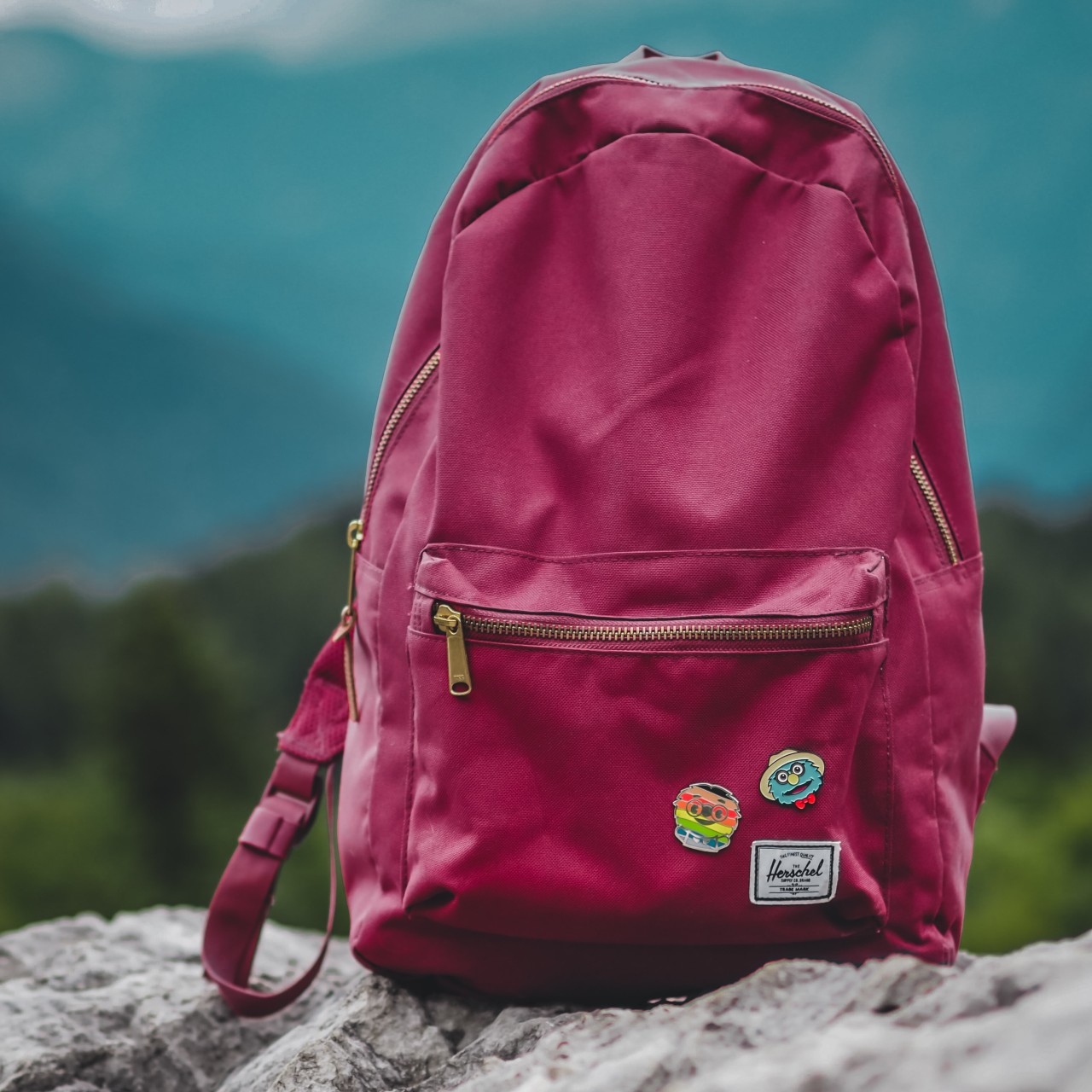}
    \includegraphics[width=0.12\linewidth,height=0.12\linewidth,trim={0cm 0cm 0cm 0cm},clip]{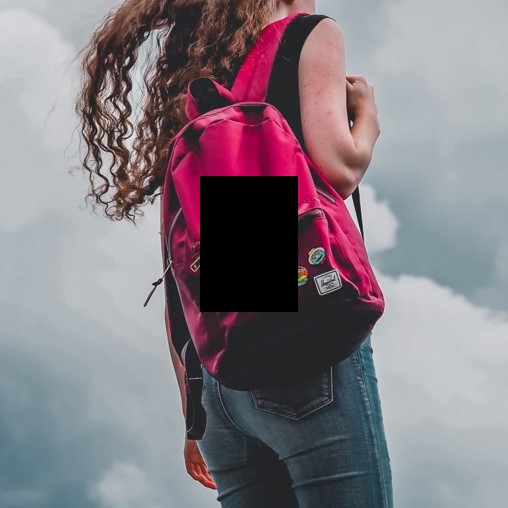}
    \includegraphics[width=0.12\linewidth,height=0.12\linewidth,trim={0cm 0cm 0cm 0cm},clip]{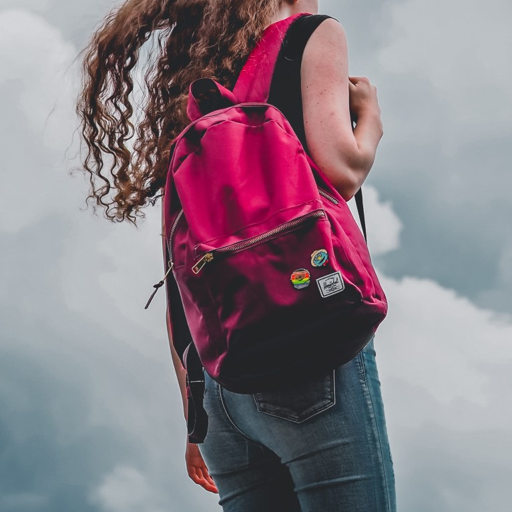}
    \hspace*{0.1em} \vline  \hspace*{0.4em}
    \includegraphics[width=0.12\linewidth,height=0.12\linewidth,trim={0cm 0cm 0cm 0cm},clip]{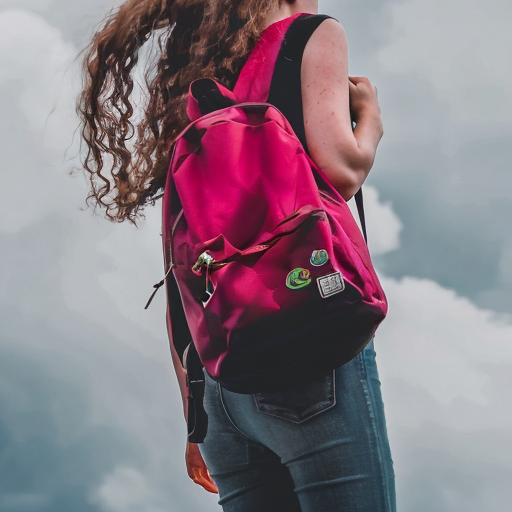}
    \includegraphics[width=0.12\linewidth,height=0.12\linewidth,trim={0cm 0cm 0cm 0cm},clip]{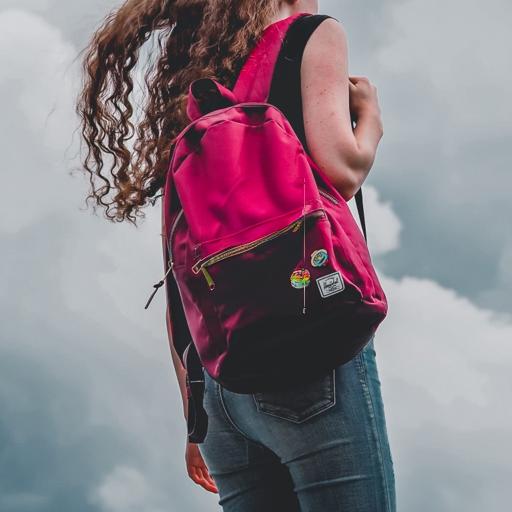}
    \includegraphics[width=0.12\linewidth,height=0.12\linewidth,trim={0cm 0cm 0cm 0cm},clip]{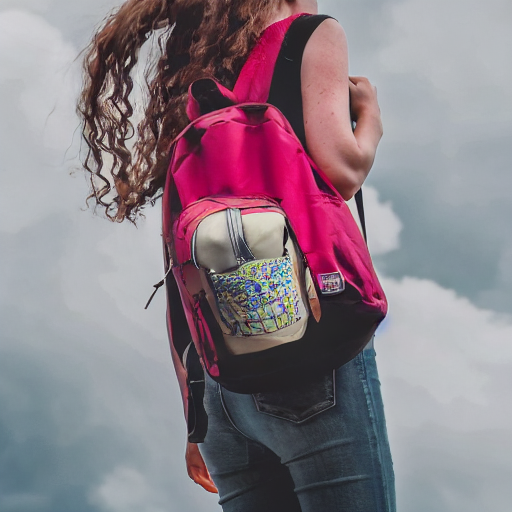}
    \includegraphics[width=0.12\linewidth,height=0.12\linewidth,trim={0cm 0cm 0cm 0cm},clip]{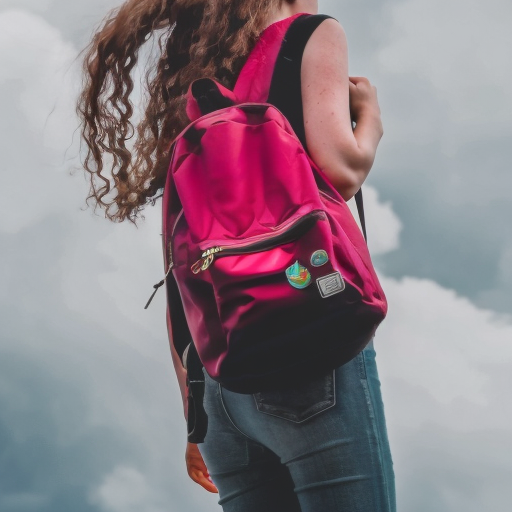} \\[0.2em]

    {\rotatebox[origin=c]{90}{\parbox{1cm}{\centering\textit{\textcolor{white}{xxx}} \\ \textit{\textcolor{white}{|}Corgi}}\hspace*{-3em}}}  
    \includegraphics[width=0.12\linewidth,height=0.12\linewidth,trim={0cm 0cm 0cm 0cm},clip]{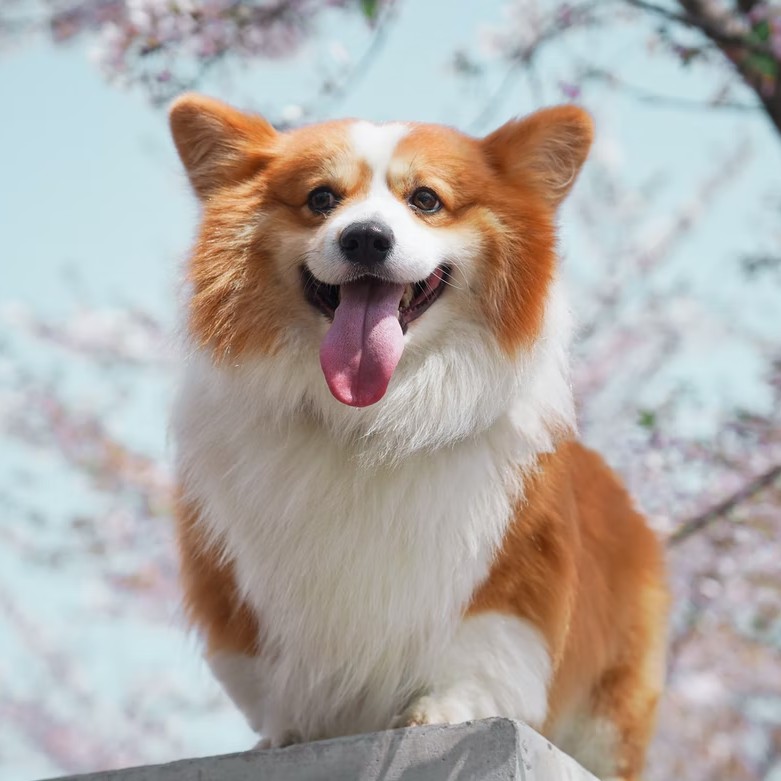} 
    \includegraphics[width=0.12\linewidth,height=0.12\linewidth,trim={0cm 0cm 0cm 0cm},clip]{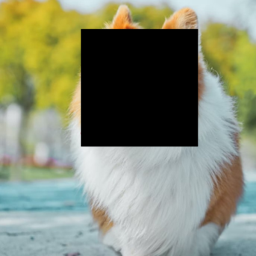}
    \includegraphics[width=0.12\linewidth,height=0.12\linewidth,trim={0cm 0cm 0cm 0cm},clip]{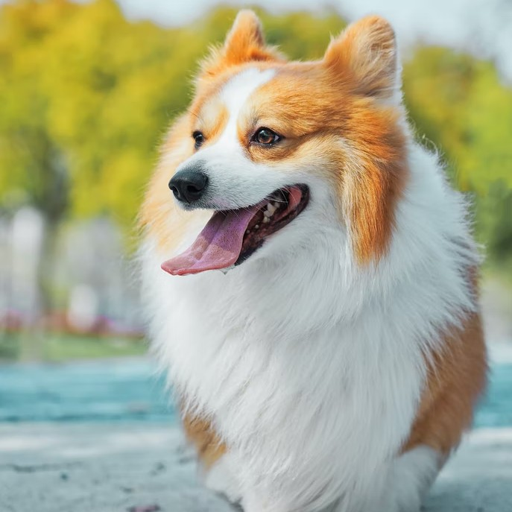}
    \hspace*{0.1em} \vline  \hspace*{0.4em}
    \includegraphics[width=0.12\linewidth,height=0.12\linewidth,trim={0cm 0cm 0cm 0cm},clip]{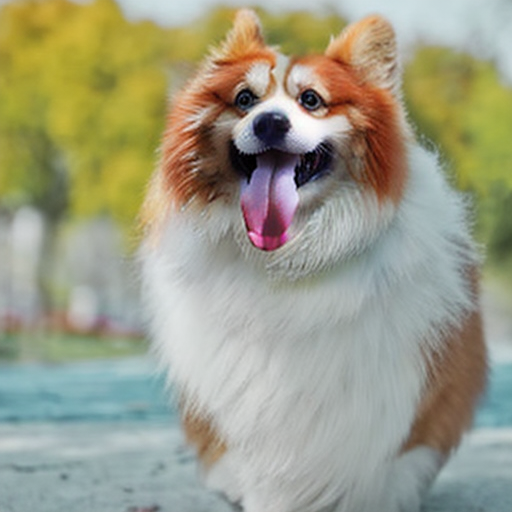}
    \includegraphics[width=0.12\linewidth,height=0.12\linewidth,trim={0cm 0cm 0cm 0cm},clip]{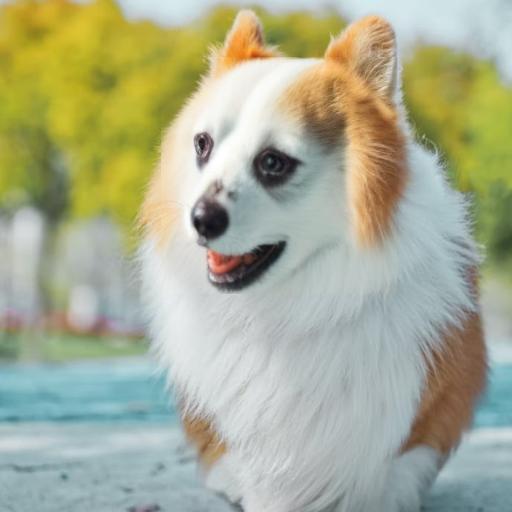}
    \includegraphics[width=0.12\linewidth,height=0.12\linewidth,trim={0cm 0cm 0cm 0cm},clip]{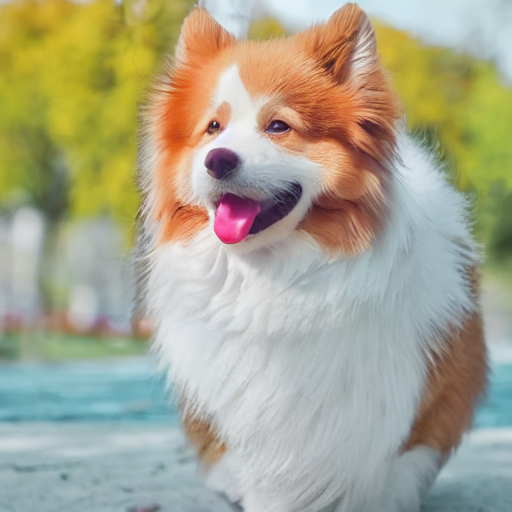}
    \includegraphics[width=0.12\linewidth,height=0.12\linewidth,trim={0cm 0cm 0cm 0cm},clip]{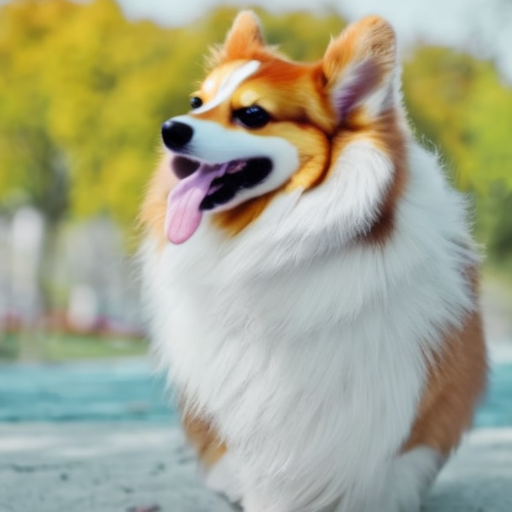} \\[0.2em]

    {\rotatebox[origin=c]{90}{\parbox{1cm}{\centering\textit{Rubber} \\ \textit{Duckie}}\hspace*{-4em}}}  
    \includegraphics[width=0.12\linewidth,height=0.12\linewidth,trim={0cm 0cm 0cm 0cm},clip]{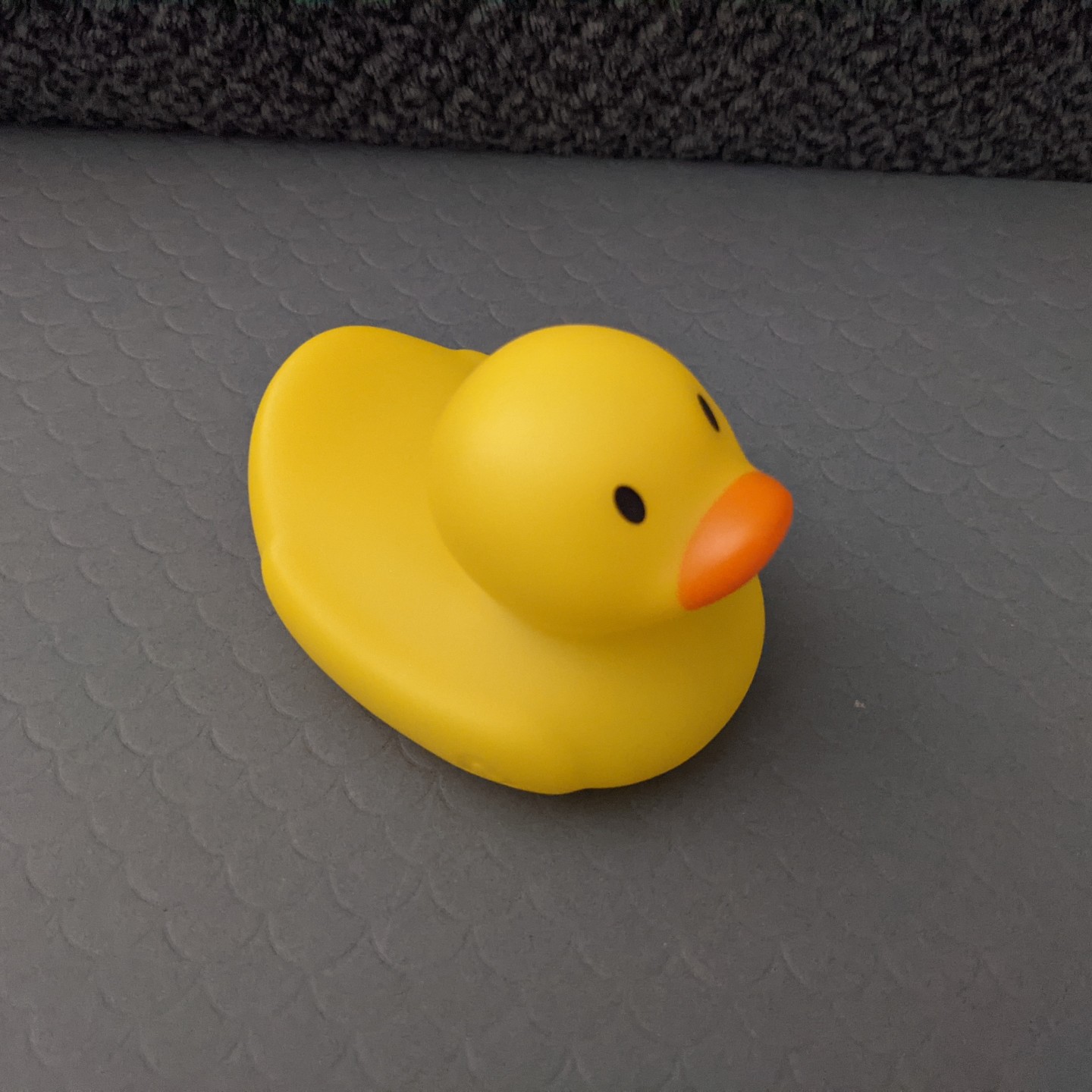}
    \includegraphics[width=0.12\linewidth,height=0.12\linewidth,trim={0cm 0cm 0cm 0cm},clip]{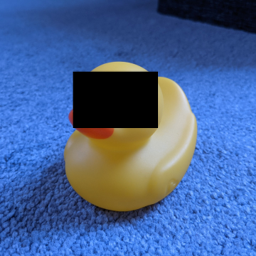}
    \includegraphics[width=0.12\linewidth,height=0.12\linewidth,trim={0cm 0cm 0cm 0cm},clip]{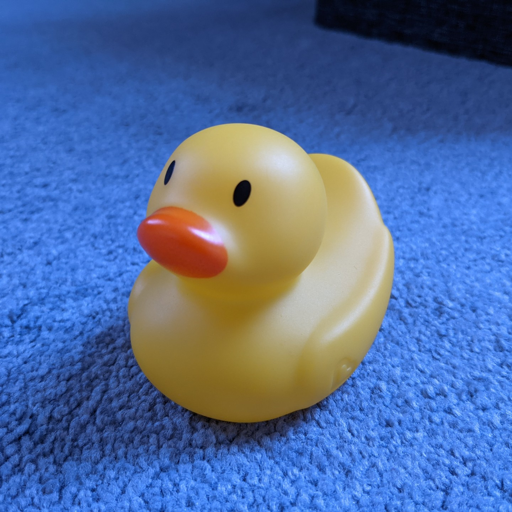}
    \hspace*{0.1em} \vline  \hspace*{0.4em}
    \includegraphics[width=0.12\linewidth,height=0.12\linewidth,trim={0cm 0cm 0cm 0cm},clip]{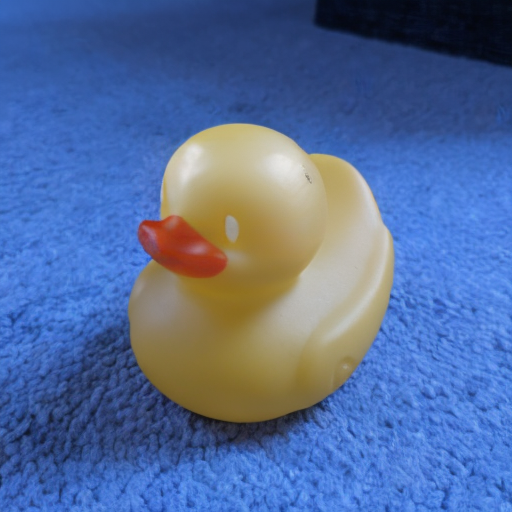}
    \includegraphics[width=0.12\linewidth,height=0.12\linewidth,trim={0cm 0cm 0cm 0cm},clip]{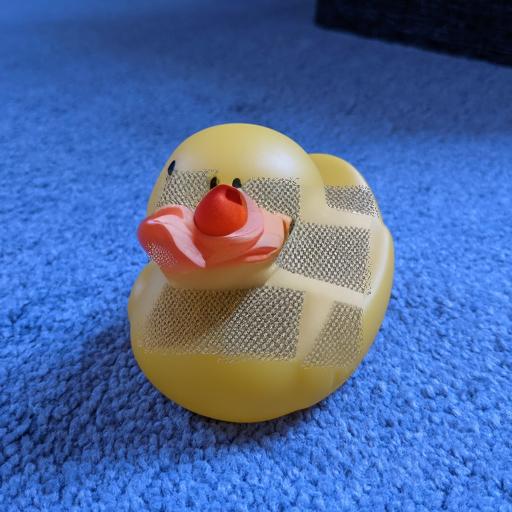}
    \includegraphics[width=0.12\linewidth,height=0.12\linewidth,trim={0cm 0cm 0cm 0cm},clip]{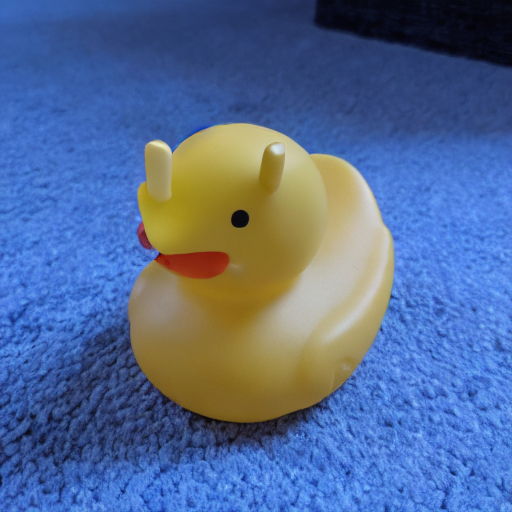}
    \includegraphics[width=0.12\linewidth,height=0.12\linewidth,trim={0cm 0cm 0cm 0cm},clip]{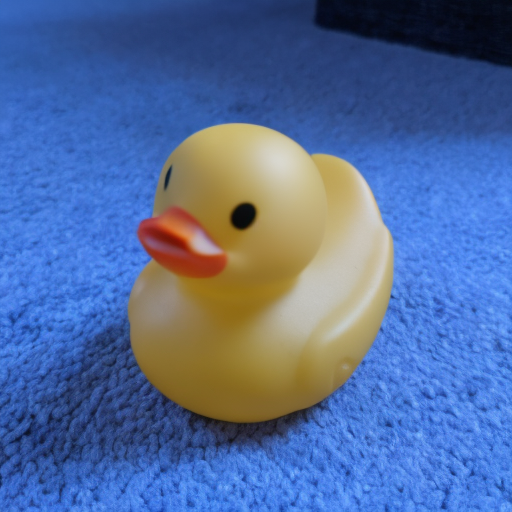} \\[0.2em]
    
    {\rotatebox[origin=c]{90}{\parbox{1cm}{\centering\textit{\textcolor{white}{xxx}} \\ \textit{\textcolor{white}{|}Projector}}\hspace*{-3em}}} 
    \includegraphics[width=0.12\linewidth,height=0.12\linewidth,trim={0cm 0cm 0cm 0cm},clip]{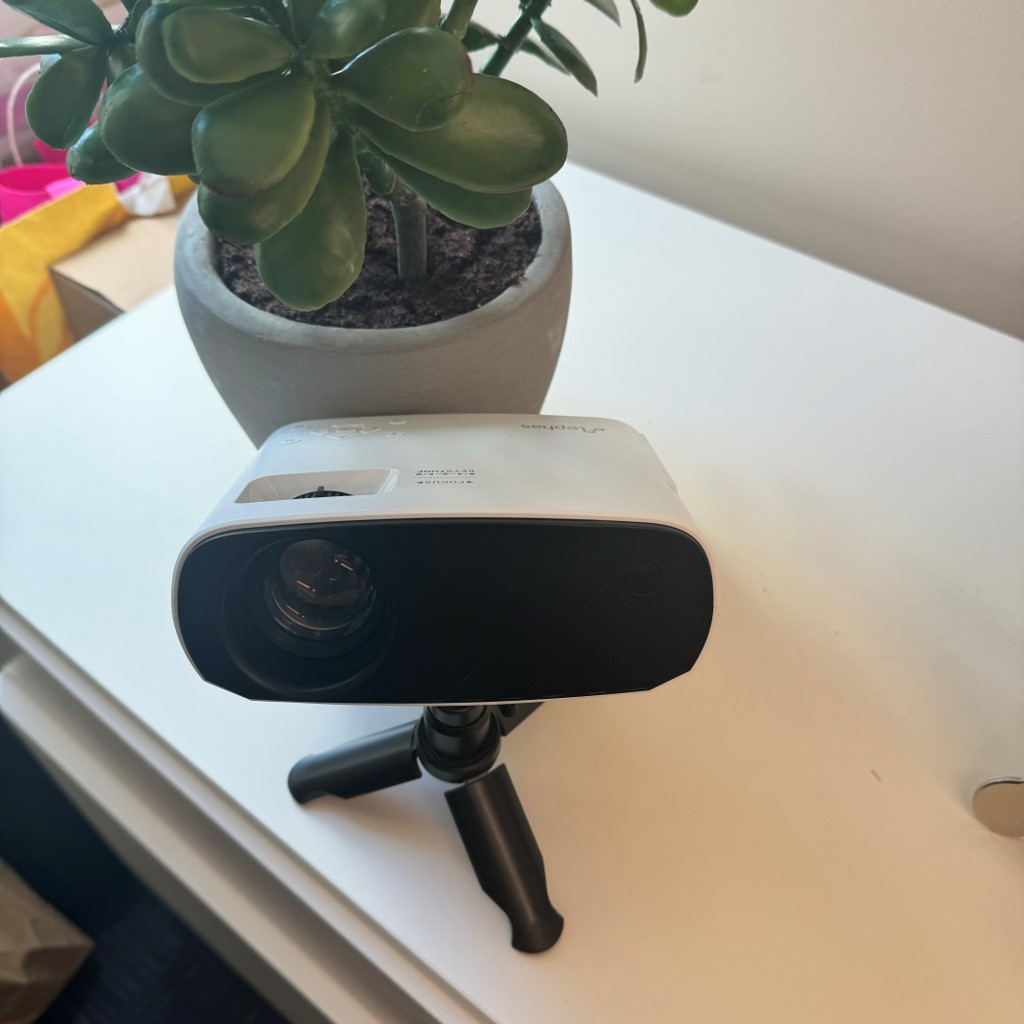} 
    \includegraphics[width=0.12\linewidth,height=0.12\linewidth,trim={0cm 0cm 0cm 0cm},clip]{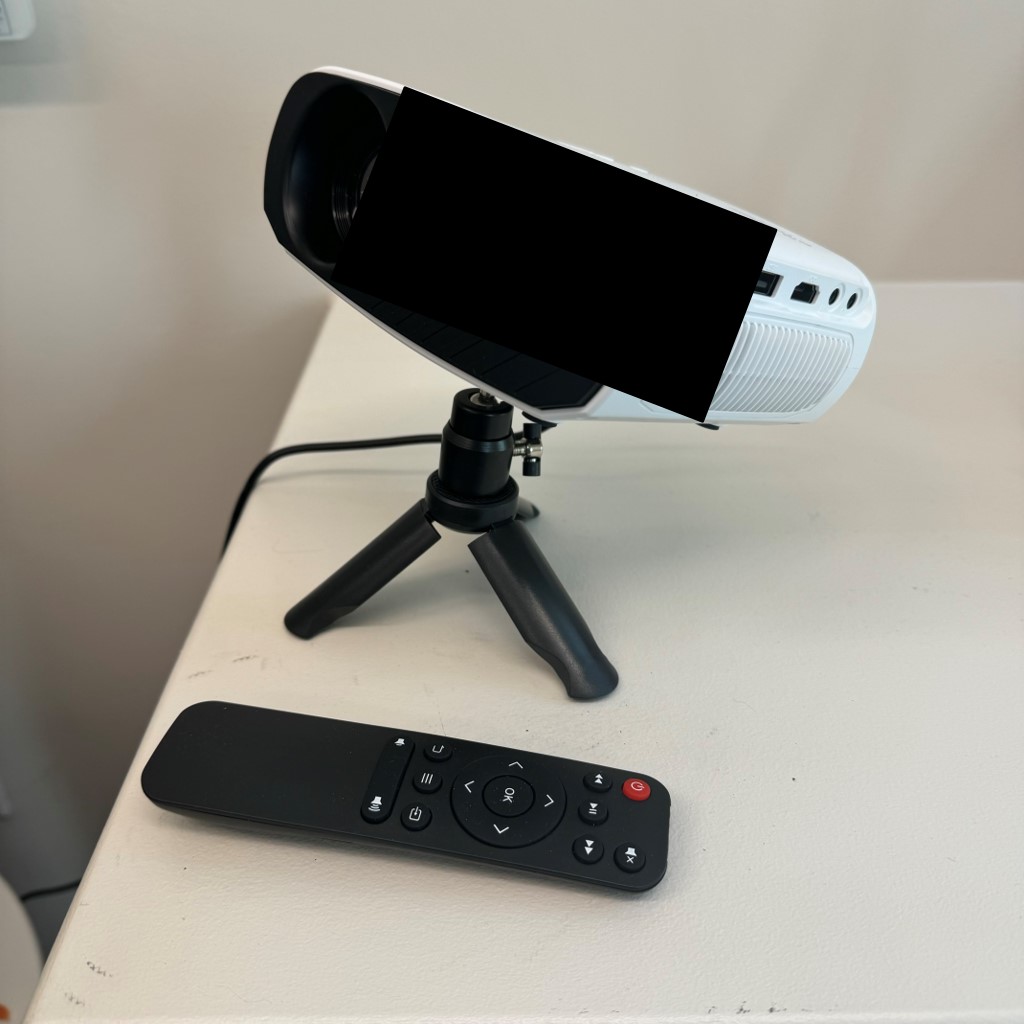}
    \includegraphics[width=0.12\linewidth,height=0.12\linewidth,trim={0cm 0cm 0cm 0cm},clip]{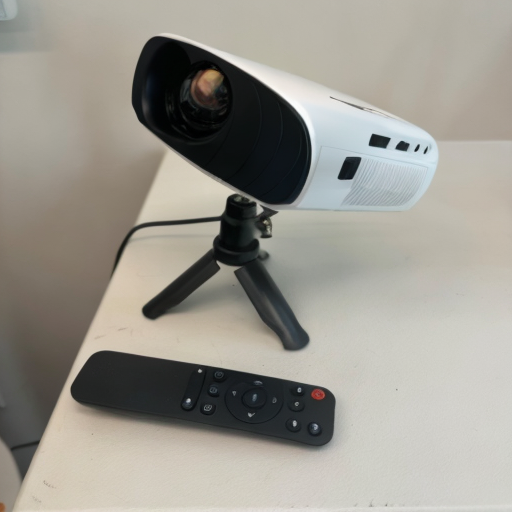} 
    \hspace*{0.1em} \vline  \hspace*{0.4em}
    \includegraphics[width=0.12\linewidth,height=0.12\linewidth,trim={0cm 0cm 0cm 0cm},clip]{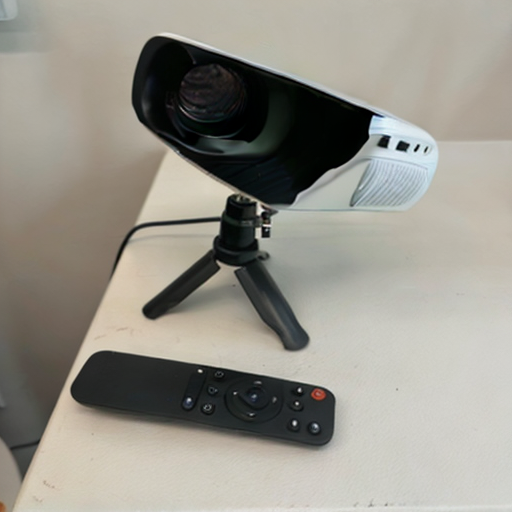}
    \includegraphics[width=0.12\linewidth,height=0.12\linewidth,trim={0cm 0cm 0cm 0cm},clip]{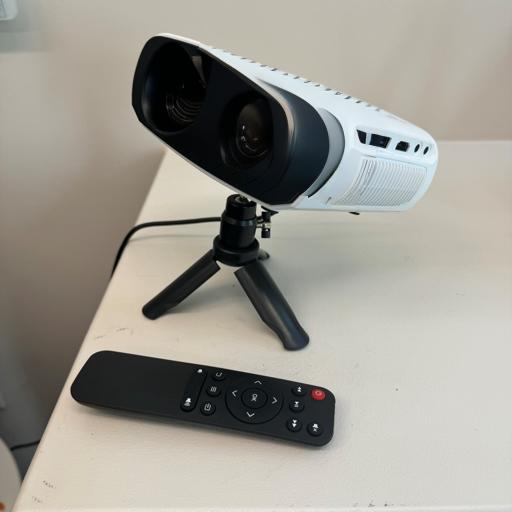}
    \includegraphics[width=0.12\linewidth,height=0.12\linewidth,trim={0cm 0cm 0cm 0cm},clip]{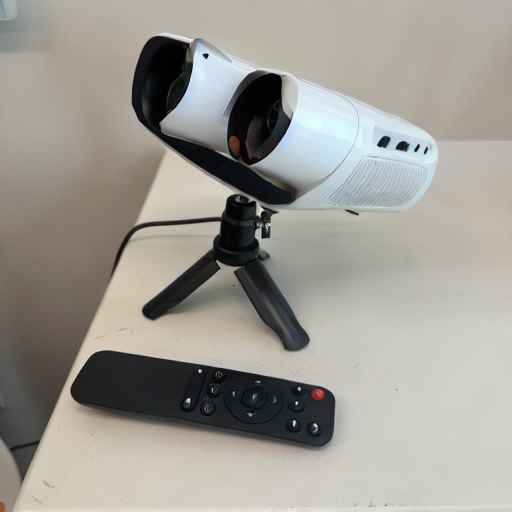}
    \includegraphics[width=0.12\linewidth,height=0.12\linewidth,trim={0cm 0cm 0cm 0cm},clip]{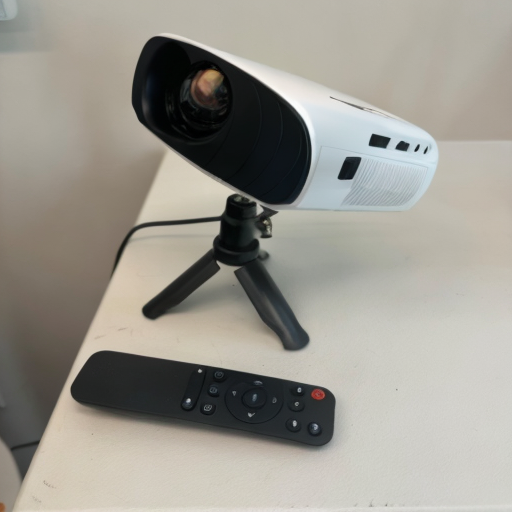} \\[0.2em]

    {\rotatebox[origin=c]{90}{\parbox{1cm}{\centering\textit{\textcolor{white}{X}} \\ \centering\textit{Suitcase}}\hspace*{-4.0em}}}   
    \includegraphics[width=0.12\linewidth,height=0.12\linewidth,trim={0cm 0cm 0cm 0cm},clip]{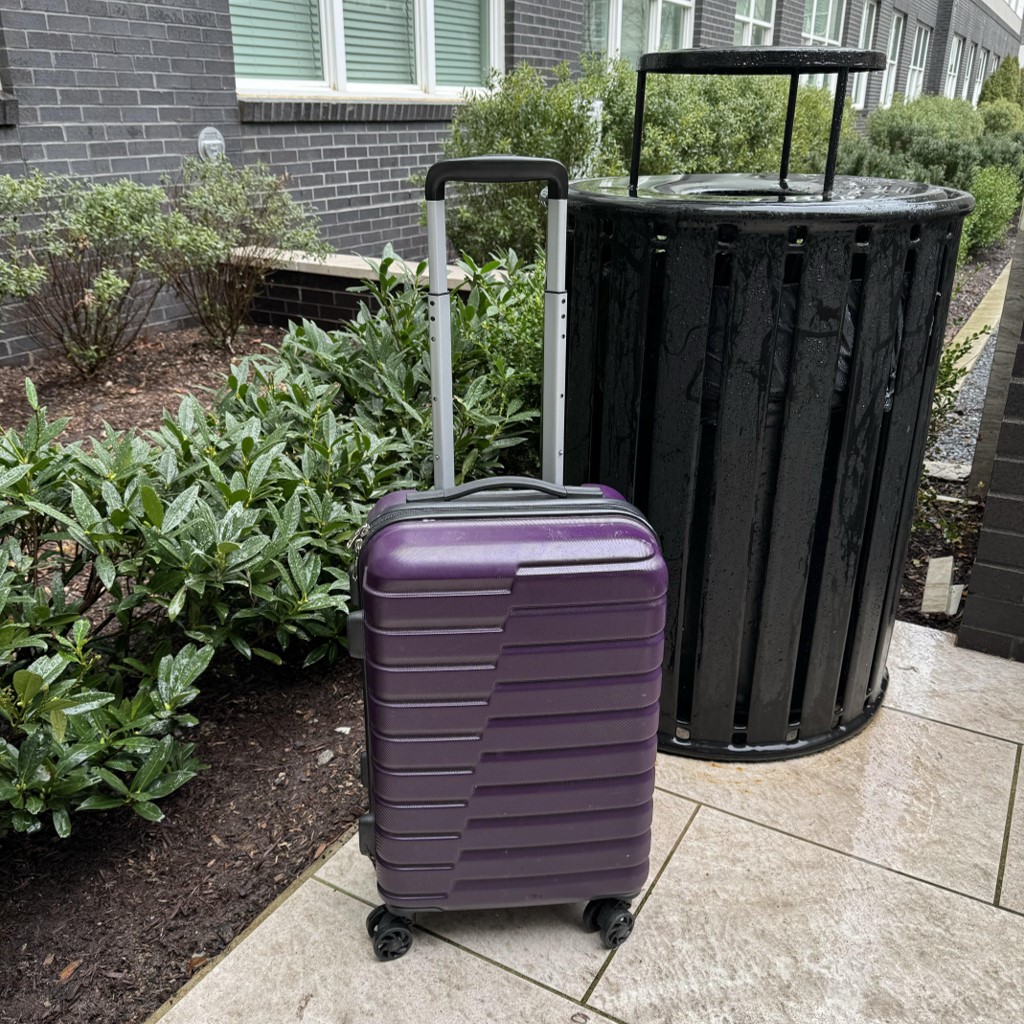} 
    \includegraphics[width=0.12\linewidth,height=0.12\linewidth,trim={0cm 0cm 0cm 0cm},clip]{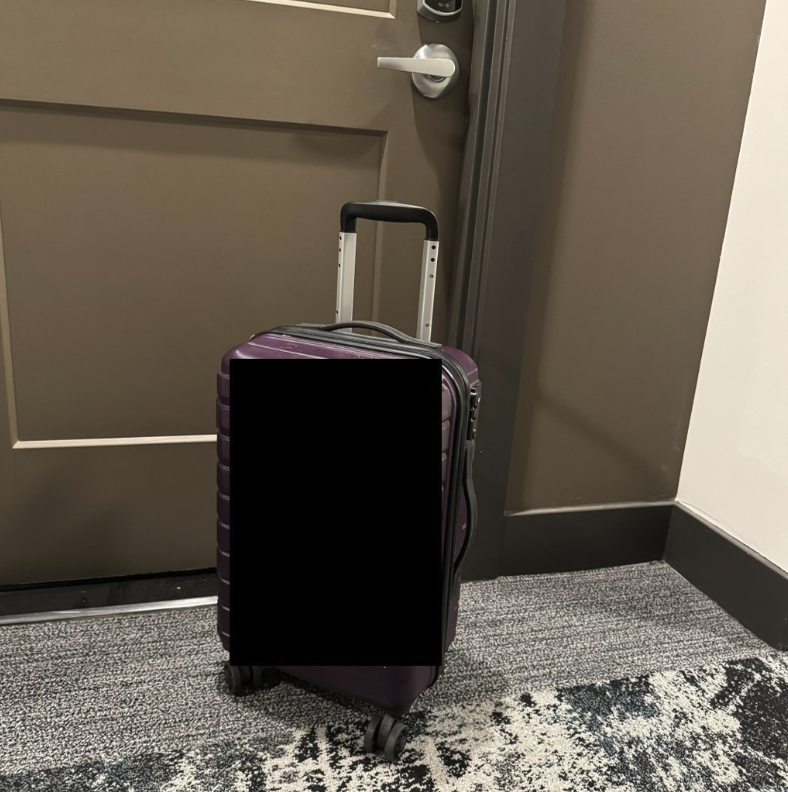}
    \includegraphics[width=0.12\linewidth,height=0.12\linewidth,trim={0cm 0cm 0cm 0cm},clip]{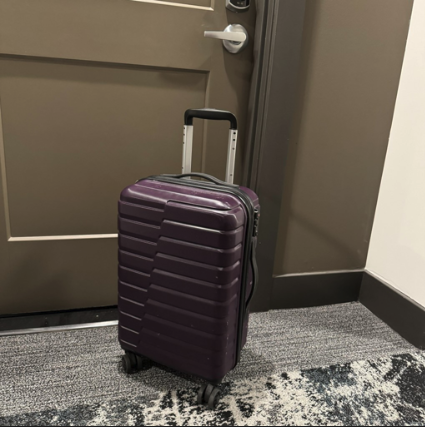} 
    \hspace*{0.1em} \vline  \hspace*{0.4em}
    \includegraphics[width=0.12\linewidth,height=0.12\linewidth,trim={0cm 0cm 0cm 0cm},clip]{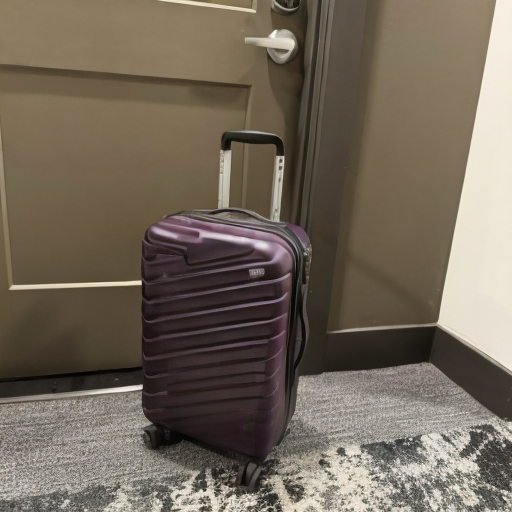}
    \includegraphics[width=0.12\linewidth,height=0.12\linewidth,trim={0cm 0cm 0cm 0cm},clip]{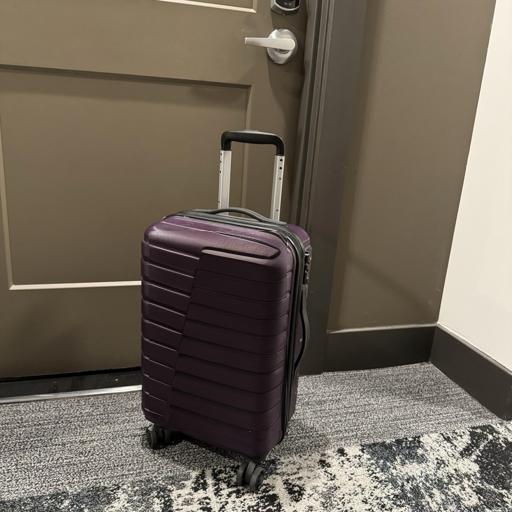}
    \includegraphics[width=0.12\linewidth,height=0.12\linewidth,trim={0cm 0cm 0cm 0cm},clip]{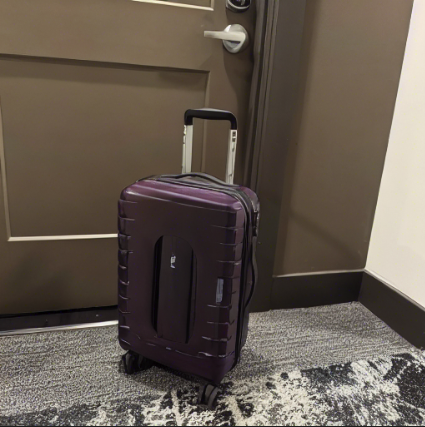}
    \includegraphics[width=0.12\linewidth,height=0.12\linewidth,trim={0cm 0cm 0cm 0cm},clip]{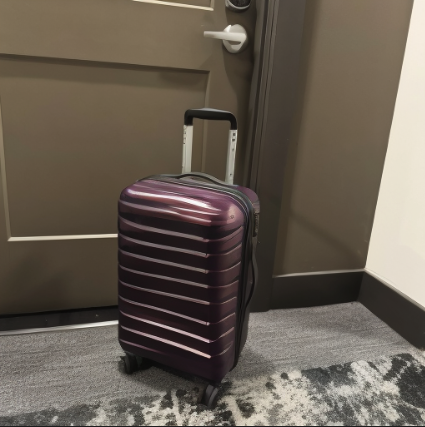} 

    {\rotatebox[origin=c]{90}{\parbox{1cm}{\centering\textit{Green} \\ \centering\textit{Wallet}}\hspace*{-4em}}}   
    \includegraphics[width=0.12\linewidth,height=0.12\linewidth,trim={0cm 0cm 0cm 0cm},clip]{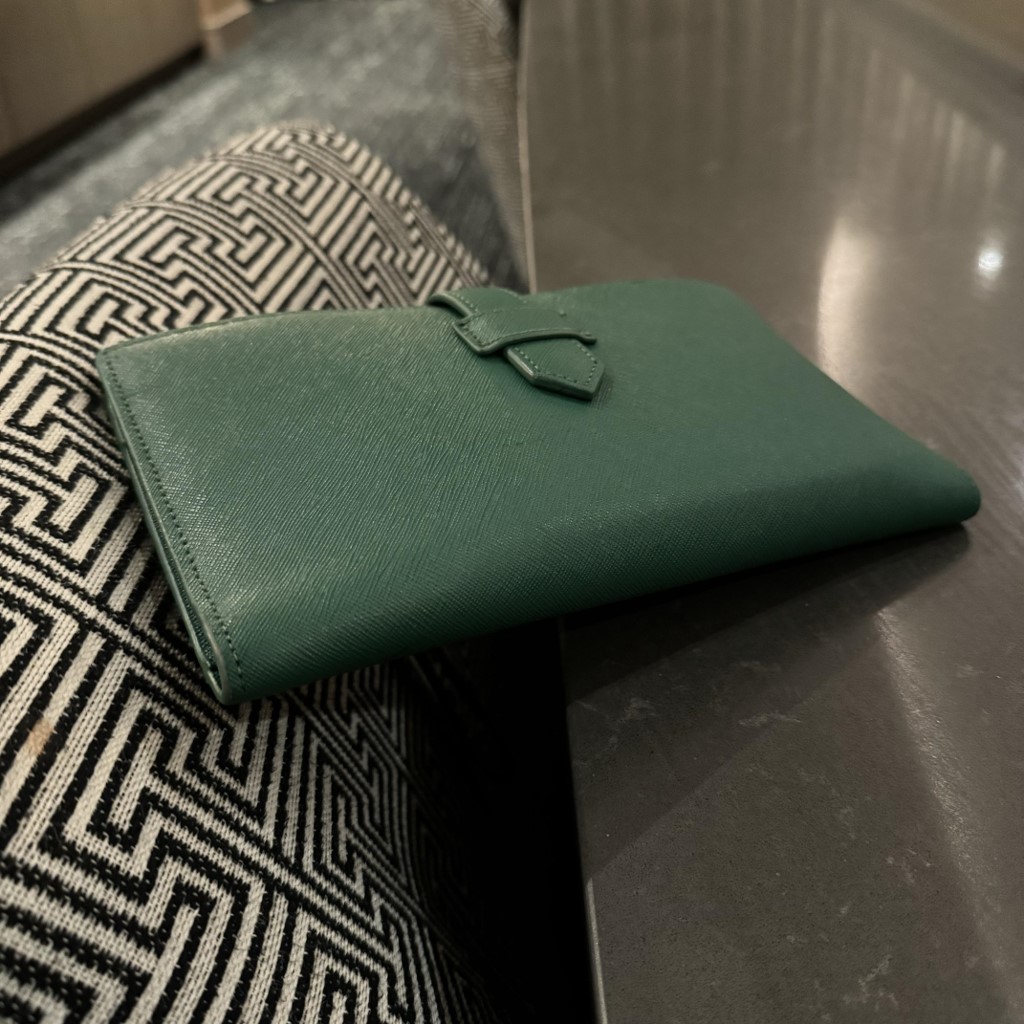} 
    \includegraphics[width=0.12\linewidth,height=0.12\linewidth,trim={0cm 0cm 0cm 0cm},clip]{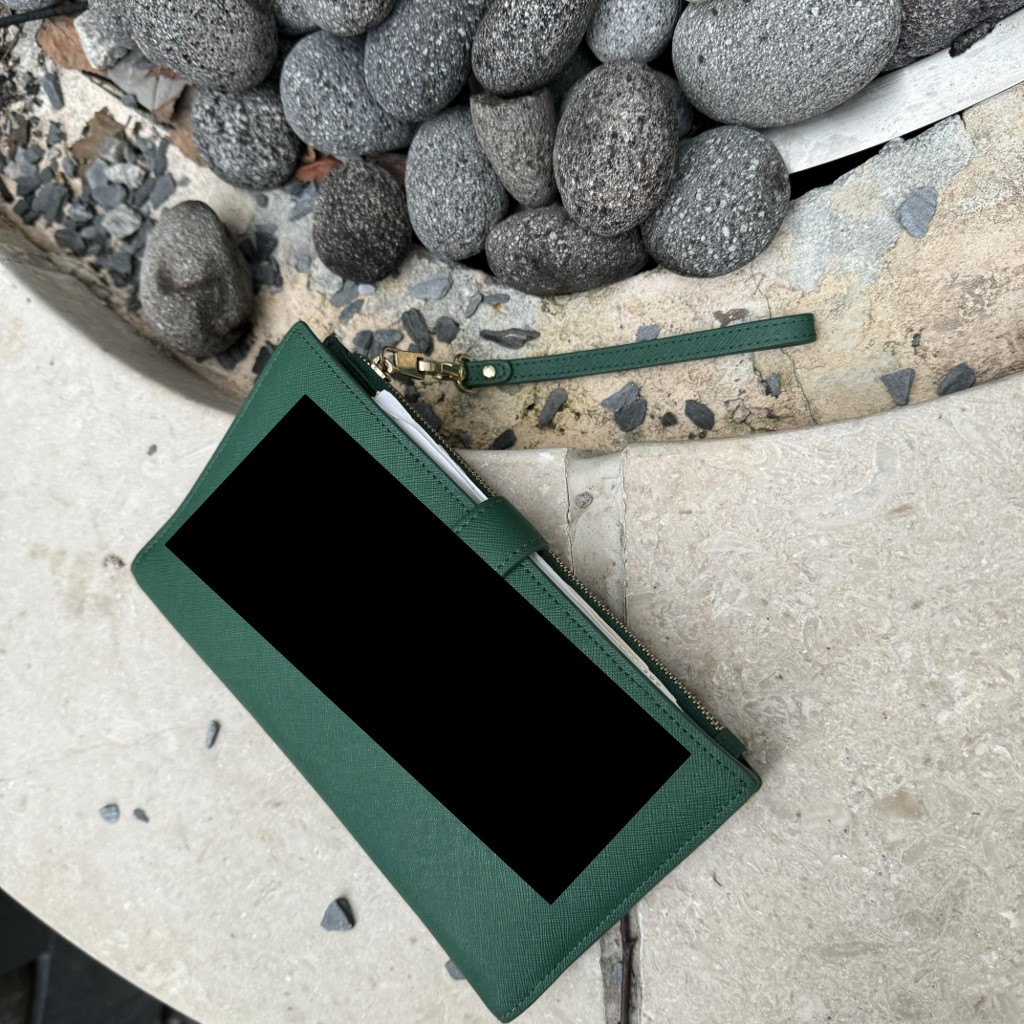}
    \includegraphics[width=0.12\linewidth,height=0.12\linewidth,trim={0cm 0cm 0cm 0cm},clip]{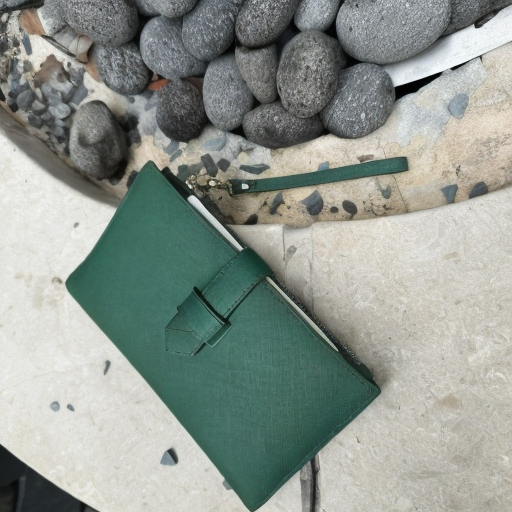} 
    \hspace*{0.1em} \vline  \hspace*{0.4em}
    \includegraphics[width=0.12\linewidth,height=0.12\linewidth,trim={0cm 0cm 0cm 0cm},clip]{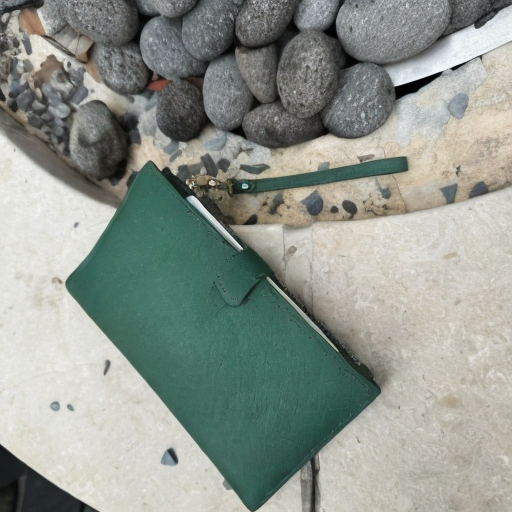}
    \includegraphics[width=0.12\linewidth,height=0.12\linewidth,trim={0cm 0cm 0cm 0cm},clip]{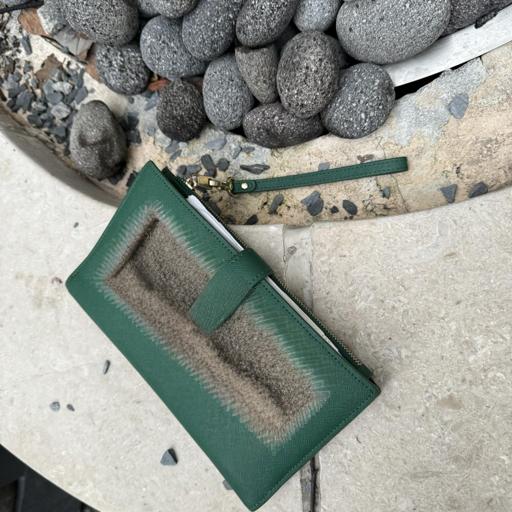}
    \includegraphics[width=0.12\linewidth,height=0.12\linewidth,trim={0cm 0cm 0cm 0cm},clip]{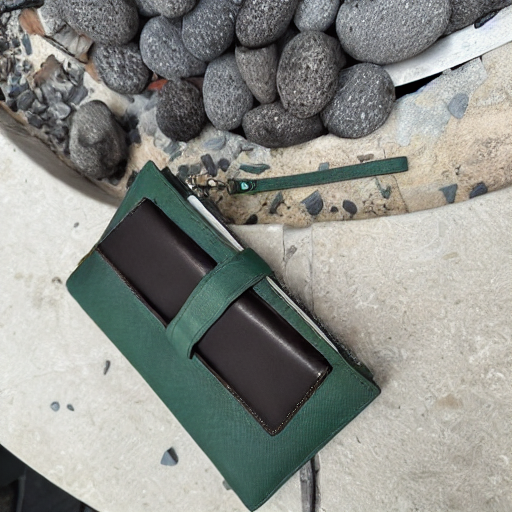}
    \includegraphics[width=0.12\linewidth,height=0.12\linewidth,trim={0cm 0cm 0cm 0cm},clip]{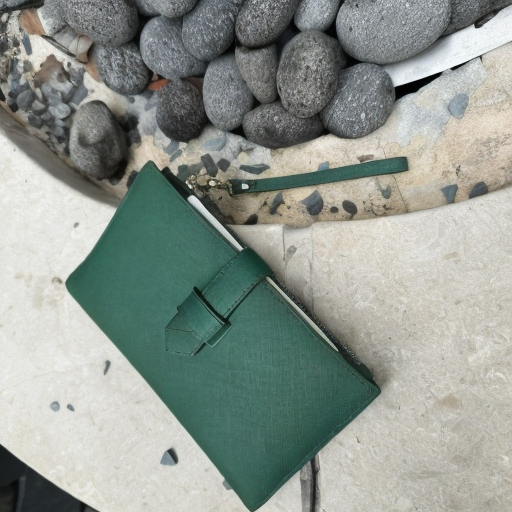}
    
    {\rotatebox[origin=c]{90}{\parbox{1cm}{\centering\textit{Monkey} \\ \centering\textit{Chair}}\hspace*{-4em}}}   
    \includegraphics[width=0.12\linewidth,height=0.12\linewidth,trim={0cm 0cm 0cm 0cm},clip]{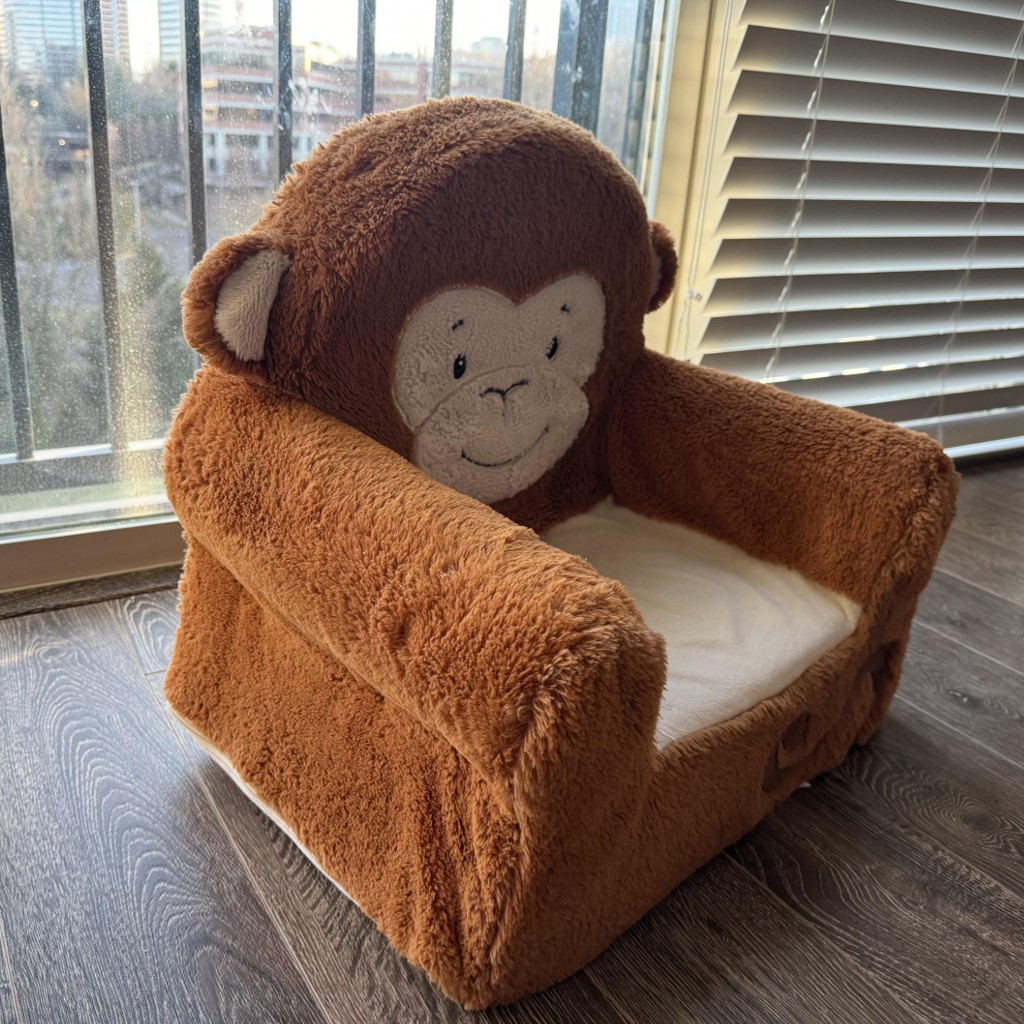} 
    \includegraphics[width=0.12\linewidth,height=0.12\linewidth,trim={0cm 0cm 0cm 0cm},clip]{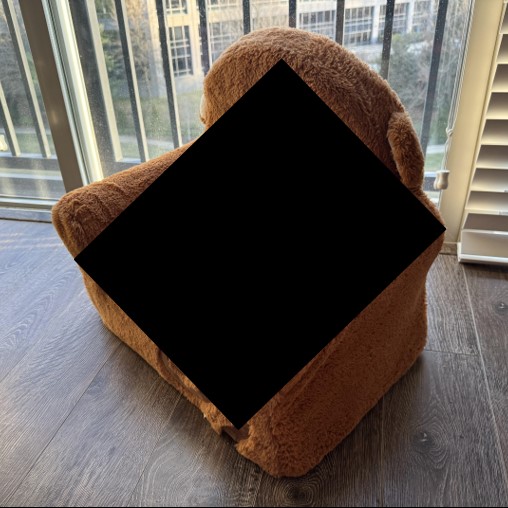}
    \includegraphics[width=0.12\linewidth,height=0.12\linewidth,trim={0cm 0cm 0cm 0cm},clip]{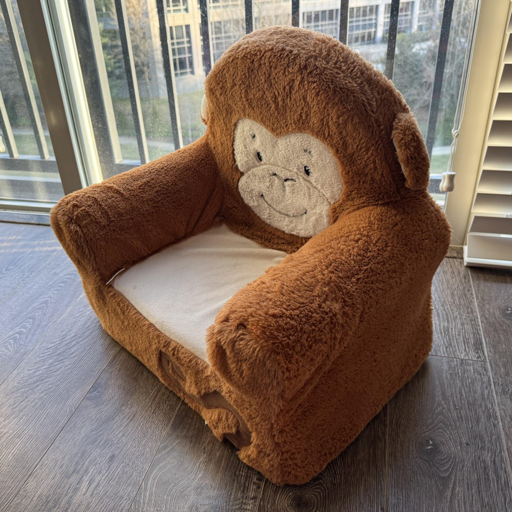} 
    \hspace*{0.1em} \vline  \hspace*{0.4em}
    \includegraphics[width=0.12\linewidth,height=0.12\linewidth,trim={0cm 0cm 0cm 0cm},clip]{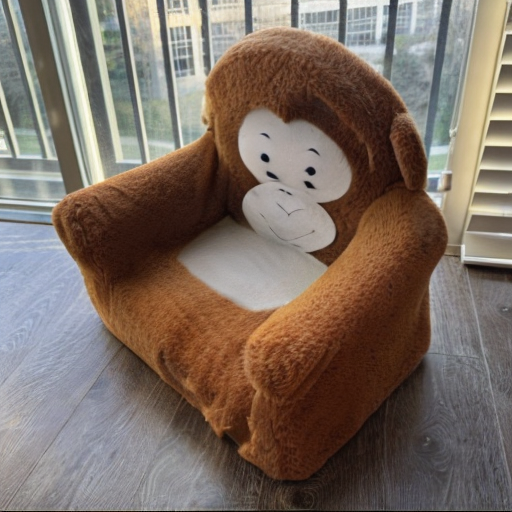}
    \includegraphics[width=0.12\linewidth,height=0.12\linewidth,trim={0cm 0cm 0cm 0cm},clip]{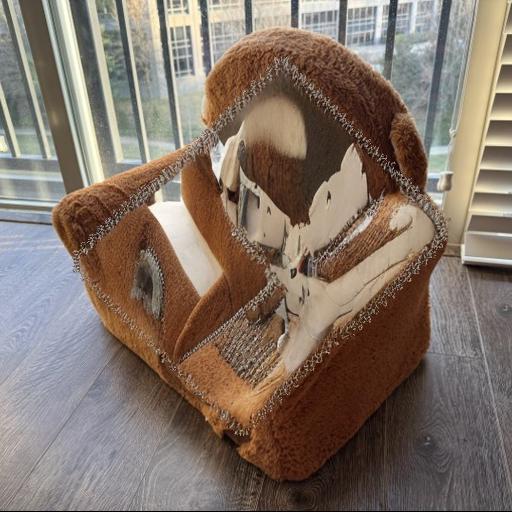}
    \includegraphics[width=0.12\linewidth,height=0.12\linewidth,trim={0cm 0cm 0cm 0cm},clip]{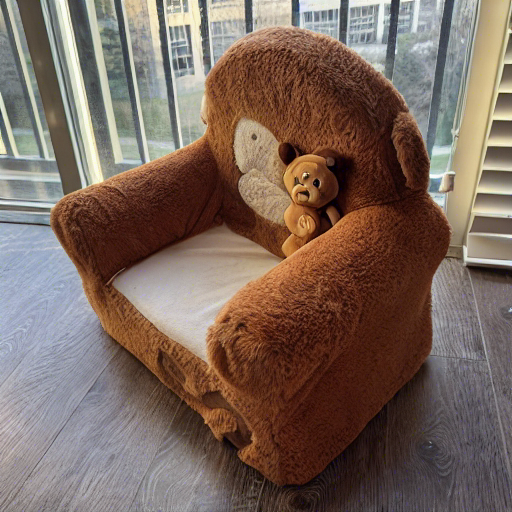}
    \includegraphics[width=0.12\linewidth,height=0.12\linewidth,trim={0cm 0cm 0cm 0cm},clip]{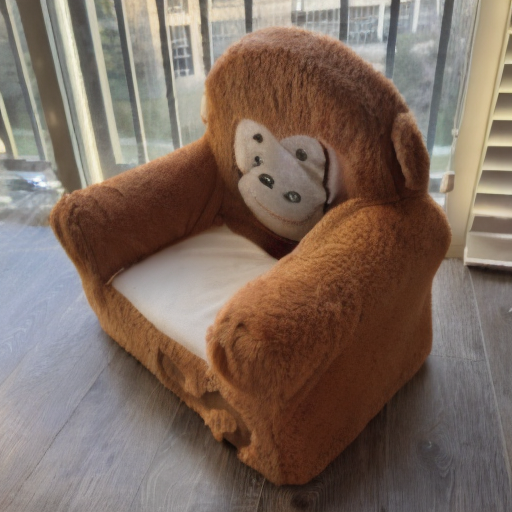}

\caption{\textbf{Qualitative Results.} Inpainting results for different missing regions using FaithFill \vs~\sota~techniques. The first four rows are images from the DreamBooth dataset, and the next four rows are images from our FaithFill dataset. Unlike Figure~\ref{fig:Intro} where we contrast state-of-the-art techniques using (or not using) a reference image, here we contrast results against the three state-of-the-art techniques that use exactly one reference image. We note that Stable Inpaining FT is an implementation of RealFill under a one-image configuration.}
\label{fig:Results}
\end{figure*}




\subsection{Results: Comparisons to State-of-the-Art}

In this work, we are comparing against seven different \sota~techniques that are diffusion based: \textit{RePaint}, \textit{GLIDE}, \textit{Blended Latent Diffusion}, \textit{Stable Inpainting}, \textit{Stable Inpainting FT}, \textit{Paint-By-Example}, and \textit{LeftRefill}. The first four do not use a reference image, and therefore have no prior for faithful reconstruction. 
In contrast, FaithFill finetunes on a single reference image. We therefore additionally compare against the three state-of-the-art methods that employ reference images for inpainting: \textit{Stable Inpainting FT} is a version of \cite{tang2023realfill} that uses a single reference image for fair comparison. \textit{Paint-By-Example} finetunes on the OpenImages dataset and uses a single reference image at inference time. In contrast, we are only finetuning on a single reference image. \textit{LeftRefill} finetunes on a single reference image to generate multiple views using an autoregressive NVS approach, while using a frozen inpainting pipeline. In contrast, we use a one-shot approach for generating the views and finetune an inpainting module on these generated  views.


\begin{table*}[t]
    \caption{\textbf{Evaluation: Image Similarity Metrics}  for all objects in the DreamBooth and FaithFill Datasets. Inpainted missing regions are evaluated with respect to the target image.}
    \centering
    \scalebox{0.92}{
    \begin{tabular}{c c|c c c | c| c c}
    \toprule
    & \textbf{Methodology} & {SSIM $\uparrow$}    & {PSNR $\uparrow$}  & {LPIPS $\downarrow$} & DreamSIM $\downarrow$ & DINO $\uparrow$ & CLIP $\uparrow$  \\
    \midrule
    \midrule
         \multirow{8}{*}{\rotatebox[origin=c]{90}{\textit{DreamBooth Dataset}\hspace*{0em}}}
         & RePaint \cite{Lugmayr_2022_CVPR} & 0.77 & 18.81 &  0.29 &  0.22 &  0.92 & 0.94 \\
         & GLIDE \cite{pmlr-v162-nichol22a} &  0.74 & 17.90 & 0.32 & 0.24 & 0.91 & 0.94  \\
         & Blended Latent Diffusion \cite{Avrahami_2022_CVPR} & 0.64 & 16.10 & 0.30 &  0.23 & 0.93 & 0.94 \\
         & Paint-By-Example \cite{yang2022paint} & 0.64 &  17.65 &  0.30 & 0.14 & 0.95  & 0.95 \\
         & Stable Inpainting \cite{Rombach_2022_CVPR_stable_diffusion} & 0.67 & 18.37 & 0.26 & 0.11 & 0.95 & 0.96  \\
         & Stable Inpainting FT \cite{Rombach_2022_CVPR_stable_diffusion, tang2023realfill} & 0.66 &  18.45 & 0.26 & 0.13 & 0.95 & 0.97  \\
         & LeftRefill \cite{cao2024leftrefill} & 0.79 & 20.84 &  0.24 &  0.12 & 0.96 & 0.96  \\
         & \textbf{FaithFill (Ours)} & 0.70 & 21.23 &  0.24  & 0.11 & 0.96 & 0.96 \\
        \midrule
         \multirow{8}{*}{\rotatebox[origin=c]{90}{\textit{FaithFill Dataset}\hspace*{0em}}}
         & RePaint \cite{Lugmayr_2022_CVPR} & 0.64 & 17.75 & 0.42 & 0.28 & 0.87 & 0.91 \\       
         & GLIDE \cite{pmlr-v162-nichol22a} & 0.64 &  16.96 & 0.40 & 0.29 & 0.88 & 0.92  \\
         & Blended Latent Diffusion \cite{Avrahami_2022_CVPR} & 0.57 & 15.76 & 0.35 & 0.29 & 0.88 & 0.93 \\
        & Paint-By-Example \cite{yang2022paint} & 0.58 & 16.90 &  0.33 &0.21 & 0.91 & 0.94 \\
         & Stable Inpainting \cite{Rombach_2022_CVPR_stable_diffusion} & 0.60 & 18.00 &  0.29 & 0.15 & 0.93 & 0.95 \\
         & Stable Inpainting FT \cite{Rombach_2022_CVPR_stable_diffusion, tang2023realfill} &  0.59 &  17.38 & 0.33 & 0.17 & 0.92 & 0.95 \\
         & LeftRefill \cite{cao2024leftrefill} & 0.65 & 19.44 & 0.33 & 0.16 & 0.93 & 0.95 \\      
         & \textbf{FaithFill (Ours)} & 0.66 & 20.15 &  0.25  & 0.11 & 0.95 & 0.97 \\
    \midrule
    \midrule
    \end{tabular}}

    \label{tab:evaluations_metrics_part_square}
\end{table*}

Figure~\ref{fig:Results} presents additional qualitative results to those presented in Figure~\ref{fig:Intro} for \sota~techniques: \textit{Stable Inpainting FT}, \textit{LeftRefill}, and \textit{Paint-By-Example}. FaithFill generated images are more faithful to the original object attributes compared to \sota~techniques that use a single reference image. Images from both the DreamBooth and FaithFill datasets are presented. 

Table~\ref{tab:evaluations_metrics_part_square} presents the quantitative evaluations of FaithFill \vs~\sota~techniques for the DreamBooth and FaithFill datasets, respectively. We note that Stable Inpainting FT represents the RealFill pipeline under a one-image configuration. In Table~\ref{tab:evaluations_metrics_part_square}, metrics are grouped in low, mid, and high level metrics as defined in ~\cite{fu2023dreamsim} by Fu \etal. 
Low level metrics (SSIM, PSNR, and LPIPS) are used to evaluate model performance on the pixel level.  DreamSIM is a mid-level similarity metric that captures differences in coarse details, such as object pose, semantic features, and image layout.  DINO and CLIP are high-level metrics that aim to asses the similarity between original and generated images using high-level feature correlation. FaithFill achieves best results on most metrics compared to other state-of-the-art methods and better or comparable results to the concurrent work LeftRefill~\cite{cao2024leftrefill}.

\begin{figure*}[t]
    \centering
    \includegraphics[width=0.95\textwidth]{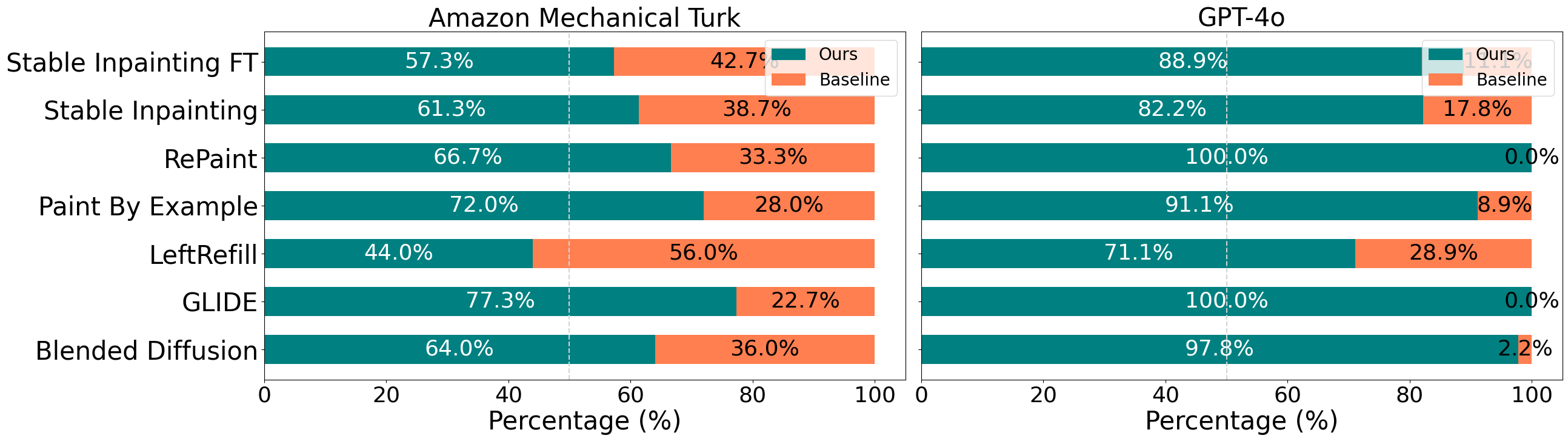}\hfill
    \caption{\textbf{Human Judgement and GPT Evaluation Results.} This figure shows the results of our user study (Left) on Amazon Mechanical Turk and GPT-4o study (Right). Every bar presents the percentage of times FaithFill generations were favored compared to the generations from state-of-the-art techniques.}
    \label{fig:gpt_study}
\end{figure*}

Figure~\ref{fig:gpt_study} presents the results of the user study and the GPT evaluation on all images of the DreamBooth and FaithFill datasets. We can see that the FaithFill generated result is selected to be more similar to the target image over state-of-the-art methods in almost all scenarios. Concurrent work LeftRefill is favored in the AMT study 12\% of the time more than FaithFill, while FaithFill is favored in the GPT study 42.2\% of the time more than LeftRefill.

\begin{figure*}[t]
    \centering 
    \setlength{\arrayrulewidth}{1.5pt}

    \rotatebox[origin=c]{0}{
    \begin{tabular}{cccc}
    \hspace*{2em}
      \parbox{1.5cm}{\centering \textit{Reference} \\ \textit{Image}} &
      \hspace*{0.8em}
      \parbox{1.5cm}{\centering \textit{Masked} \\ \textit{Target}} &
      \hspace*{0.8em}
      \parbox{1.5cm}{\centering \textit{Target} \\ \textit{Image}} &
      \hspace*{0.8em}
      \parbox{1.5cm}{\centering \textit{\textbf{FaithFill}} \\ \textit{\textbf{(Ours)}}}
    \end{tabular}
  }

    {\rotatebox[origin=c]{90}{\parbox{1cm}{\centering\textit{Partial Occ.}}\hspace*{-4em}}} 
    \includegraphics[width=0.14\linewidth,height=0.14\linewidth,trim={0cm 0cm 0cm 0cm},clip]{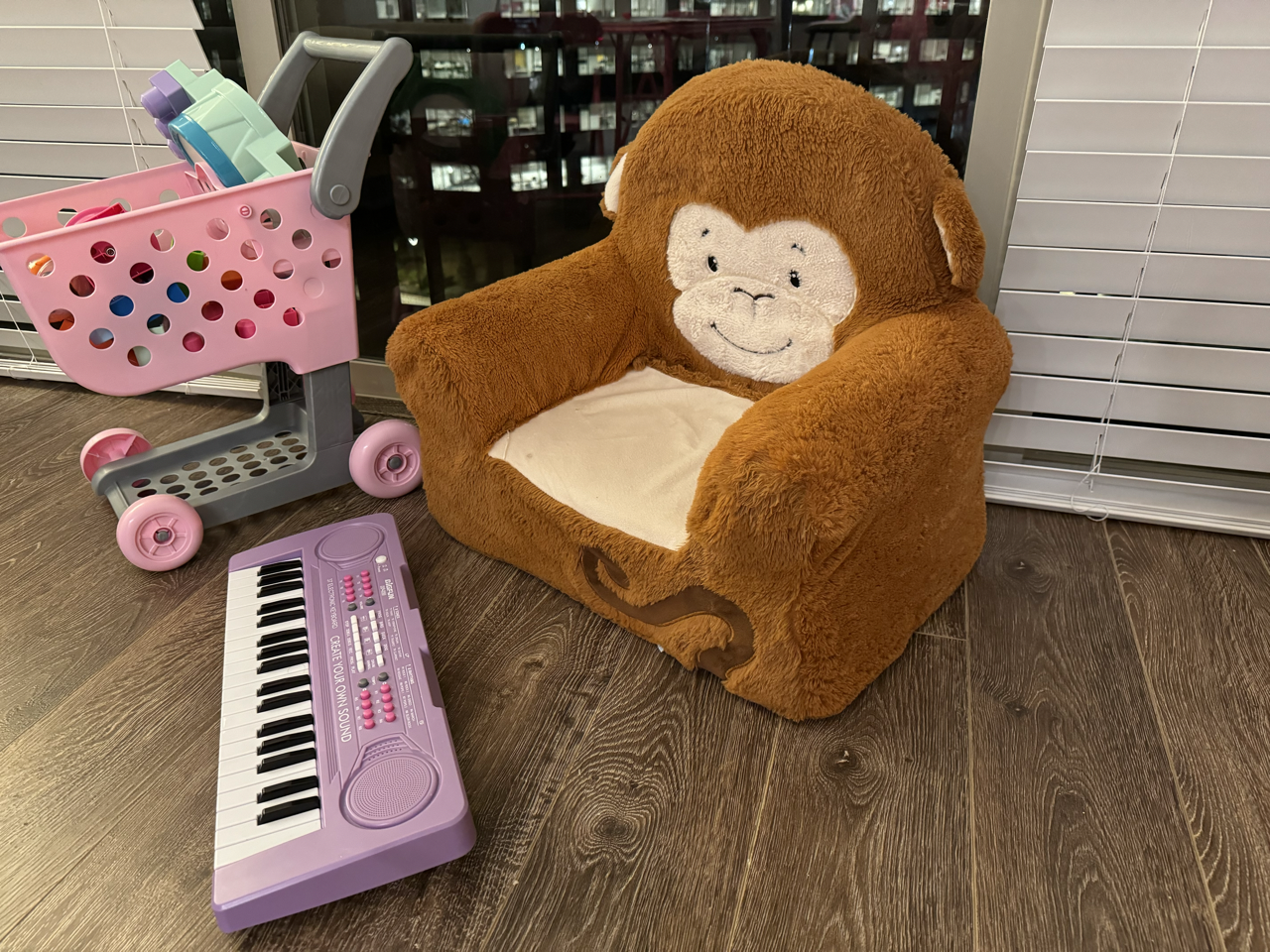} 
    \includegraphics[width=0.14\linewidth,height=0.14\linewidth,trim={0cm 0cm 0cm 0cm},clip]{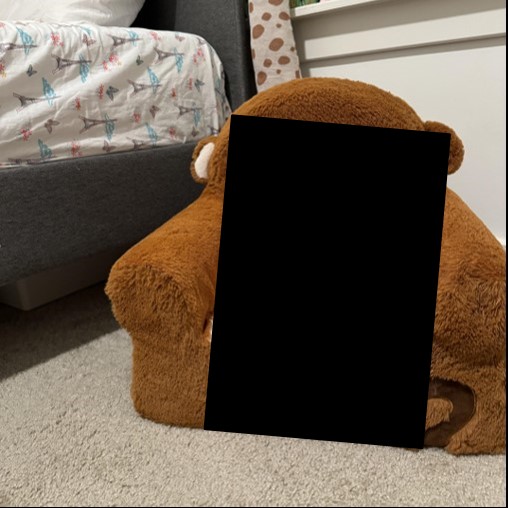}
    \includegraphics[width=0.14\linewidth,height=0.14\linewidth,trim={0cm 0cm 0cm 0cm},clip]{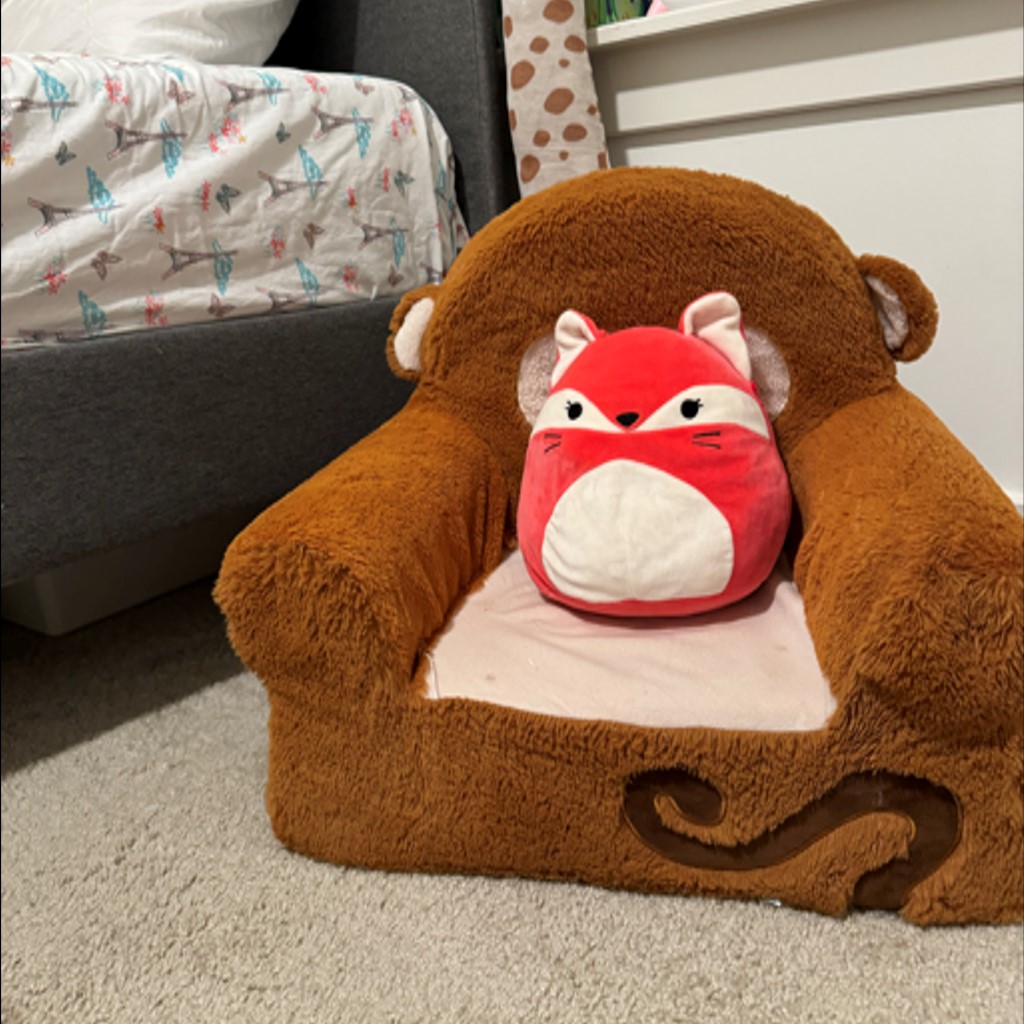} 
    \includegraphics[width=0.14\linewidth,height=0.14\linewidth,trim={0cm 0cm 0cm 0cm},clip]{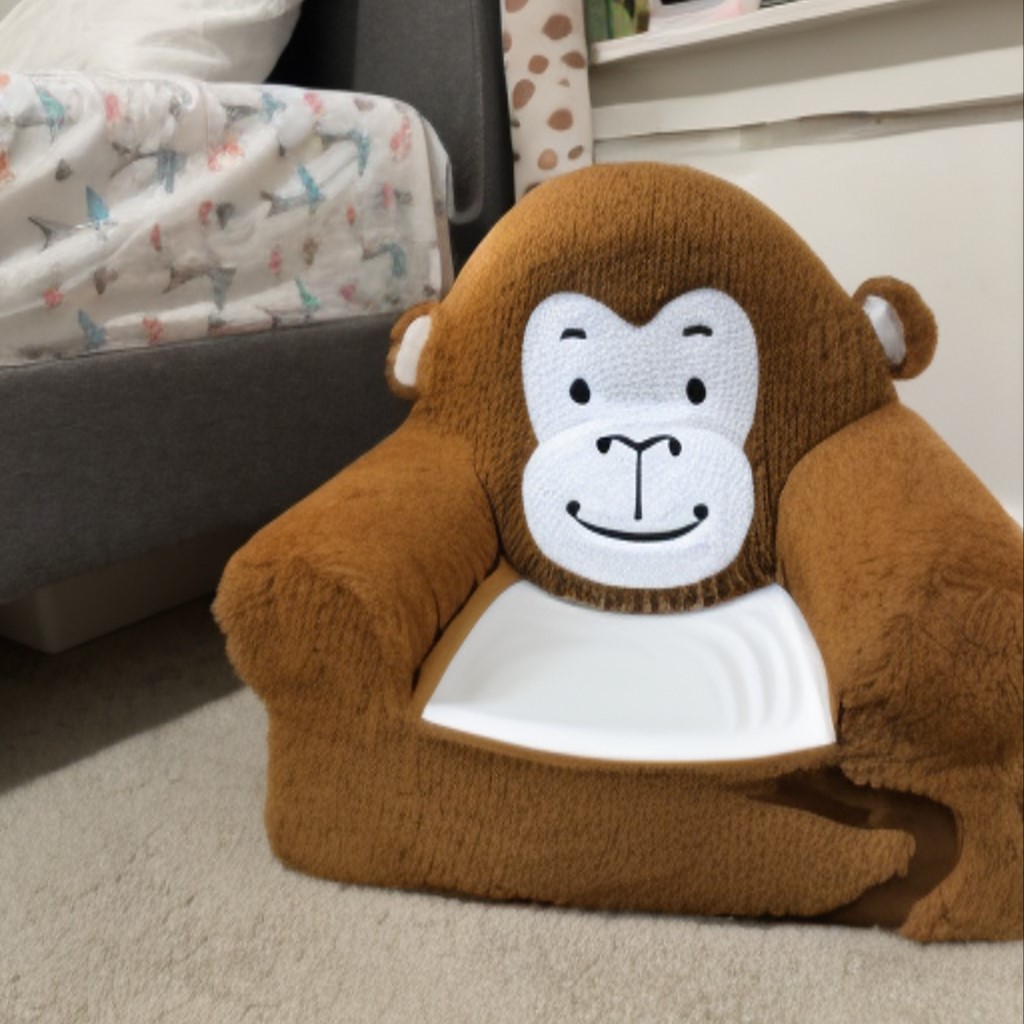} 
    
    {\rotatebox[origin=c]{90}{\parbox{1cm}{\centering\textit{Full Occ.}}\hspace*{-5em}}} 
    \includegraphics[width=0.14\linewidth,height=0.14\linewidth,trim={0cm 0cm 0cm 0cm},clip]{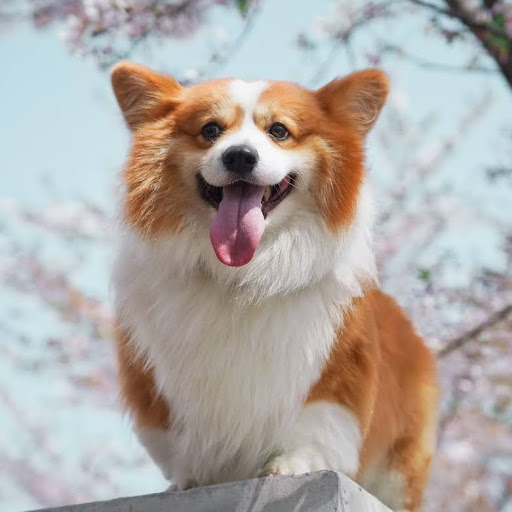} 
    \includegraphics[width=0.14\linewidth,height=0.14\linewidth,trim={0cm 0cm 0cm 0cm},clip]{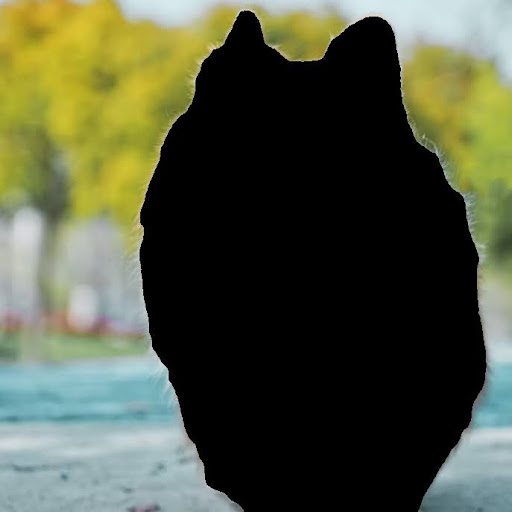}
    \includegraphics[width=0.14\linewidth,height=0.14\linewidth,trim={0cm 0cm 0cm 0cm},clip]{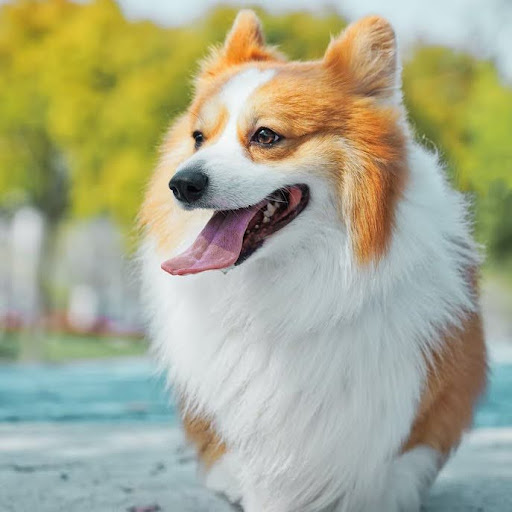} 
    \includegraphics[width=0.14\linewidth,height=0.14\linewidth,trim={0cm 0cm 0cm 0cm},clip]{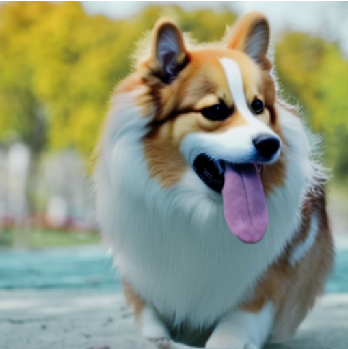} 
    
\caption{\textbf{Additional Capabilities.} This demonstrates FaithFill's ability to remove objects partially occluding the target object \textit{(row 1)}, as well as its ability to hallucinate a fully occluded object in the target image \textit{(row 2)}, based on a \textit{single} reference image.}
\label{fig:Additional_Capabilities}
\end{figure*}

Additional Capabilities of FaithFill are demonstrated in Figure~\ref{fig:Additional_Capabilities}. Such capabilities can be used for various image editing tasks, \eg~to remove objects occluding other objects while inpainting the occluded image faithfully, or to inpaint objects on different backgrounds. The former can be achieved by masking the occluding object before inference. The latter can be achieved by masking the region the object should be inserted before inference.

\section{Conclusions}
We propose a novel finetuning framework, FaithFill, for object inpainting. FaithFill completes the part of an object that is missing due to any reason from the image, \eg~occluded by another object. We focus on completing the missing region in a faithful manner, that is, maintaining the same object structure, color, and texture, together with not altering the target background. FaithFill only needs  a single reference image as input to achieve faithful inpainting of a target image. The FaithFill training pipeline includes a segmentation module to extract the object from the reference image, a diffusion based view generation module to generate multiple views of the extracted object, followed by an inpainting module that reconstructs the masked image. This work demonstrates the ability to finetune using a single reference image to produce high quality inpainting faithful to the provided reference image features, despite reasonable discrepancies in viewpoints, poses, lighting conditions, and backgrounds. Additionally, we propose the FaithFill dataset, which consists of image pairs of different objects taken in various conditions, \eg~different viewpoints, lighting, setting, \etc  to further enrich the evaluation and share with the research community.

\section{Limitations and Negative Impacts}
\textbf{Limitations. }\label{subsec:limitations}
Since FaithFill relies on NeRFs to generate multiple views, the inpainting process might become challenging when working with a reference image that is hard to generate views for (\ie~poor views implies poor inpainting).
In addition, if the viewpoints of the reference and target images are drastically different (\eg~back and front), \textit{FaithFill} will obviously struggle to inpaint the target image. Furthermore, since FaithFill depends on the prior knowledge of the pretrained base model - Stable Diffusion - it inherits some of the challenges it faces with generation of fine details, such as text. Although FaithFill is capable of hallucinating the details of a \textbf{fully} occluded object based on the reference image, it cannot guarantee generating the same pose of the original object/subject in the ground truth target image. Finally, defining the number of iterations needed for finrtuning can be tricky. In our work we set a fixed number of iterations for each dataset. Having a `slider' for each image being edited would probably give better results.

\textbf{Negative Impacts. }\label{subsec:negative_impacts}
FaithFill is a tool that can be used by members of the society to unleash their creativity and improve the quality of their images and/or photographs. Nonetheless, as inpainting is an image editing technique, this research inherits all the potential negative impacts associated with image editing, \eg~placing a political personnel  in a contentious location or alongside a controversial figure. In addition, this research involves the use of computationally intensive models, suggesting elevated energy consumption that may have environmental implications. In efforts to minimize that, we used frozen weights whenever possible. 

\bibliographystyle{splncs04}
\bibliography{references}

\end{document}